\documentclass{article}

\usepackage{microtype}
\usepackage{graphicx}
\usepackage{booktabs} 
\usepackage[table]{xcolor}
\usepackage{pifont}
\usepackage{multirow}
\usepackage{algorithm}
\usepackage{algorithmic}
\usepackage[caption=false]{subfig} 
\usepackage{amsmath}
\usepackage{amssymb}
\usepackage{mathtools}
\usepackage{amsthm}

\usepackage{hyperref}

\usepackage[accepted]{icml2025}

\definecolor{ta}{HTML}{cc7370}
\definecolor{isoc}{HTML}{2d9666}
\definecolor{isocts}{HTML}{d9a941}

\usepackage[capitalize,noabbrev]{cleveref}

\theoremstyle{plain}

\theoremstyle{definition}

\theoremstyle{remark}

\newcommand{\minisection}[1]{\vspace{0.0in}\noindent{\bf #1}\hspace{0.15em}}

\icmltitlerunning{No Task Left Behind: Isotropic Model Merging with Common and Task-Specific Subspaces}

\begin{document}

\twocolumn[
\icmltitle{No Task Left Behind:\\Isotropic Model Merging with Common and Task-Specific Subspaces}



\icmlsetsymbol{equal}{*}

\begin{icmlauthorlist}
\icmlauthor{Daniel Marczak}{pw,ideas}
\icmlauthor{Simone Magistri}{uof}
\icmlauthor{Sebastian Cygert}{nask,pg}\\
\icmlauthor{Bartłomiej Twardowski}{ideas_rc,cvc,uab}
\icmlauthor{Andrew D. Bagdanov}{uof}
\icmlauthor{Joost van de Weijer}{cvc,uab}

\end{icmlauthorlist}

\icmlaffiliation{pw}{Warsaw University of Technology, Poland}
\icmlaffiliation{ideas}{IDEAS NCBR, Warsaw, Poland}
\icmlaffiliation{ideas_rc}{IDEAS Research Center, Warsaw, Poland}
\icmlaffiliation{uof}{Department of Information Engineering, Media Integration and Communication Center (MICC), University of Florence, Italy}
\icmlaffiliation{nask}{NASK - PIB, National Research Institute, Warsaw, Poland}
\icmlaffiliation{pg}{Gdańsk University of Technology, Poland}
\icmlaffiliation{cvc}{Computer Vision Center, Barcelona, Spain}
\icmlaffiliation{uab}{Department of Computer Science, Universitat Autonoma de Barcelona, Spain}

\icmlcorrespondingauthor{Daniel Marczak}{daniel.marczak.dokt@pw.edu.pl}

\vskip 0.3in
]



\printAffiliationsAndNotice{}  

\begin{abstract}
Model merging integrates the weights of multiple task-specific models into a single multi-task model. Despite recent interest in the problem, a significant performance gap between the combined and single-task models remains.  In this paper, we investigate the key characteristics of task matrices -- weight update matrices applied to a pre-trained model -- that enable effective merging. We show that alignment between singular components of task-specific and merged matrices strongly correlates with performance improvement over the pre-trained model. Based on this, we propose an isotropic merging framework that flattens the singular value spectrum of task matrices, enhances alignment, and reduces the performance gap. Additionally, we incorporate both common and task-specific subspaces to further improve alignment and performance. Our proposed approach achieves state-of-the-art performance on vision and language tasks across various sets of tasks and model scales. This work advances the understanding of model merging dynamics, offering an effective methodology to merge models without requiring additional training.
\end{abstract}

\section{Introduction}

\begin{figure}[t!]
    \centering
    \includegraphics[width=1.0\linewidth]{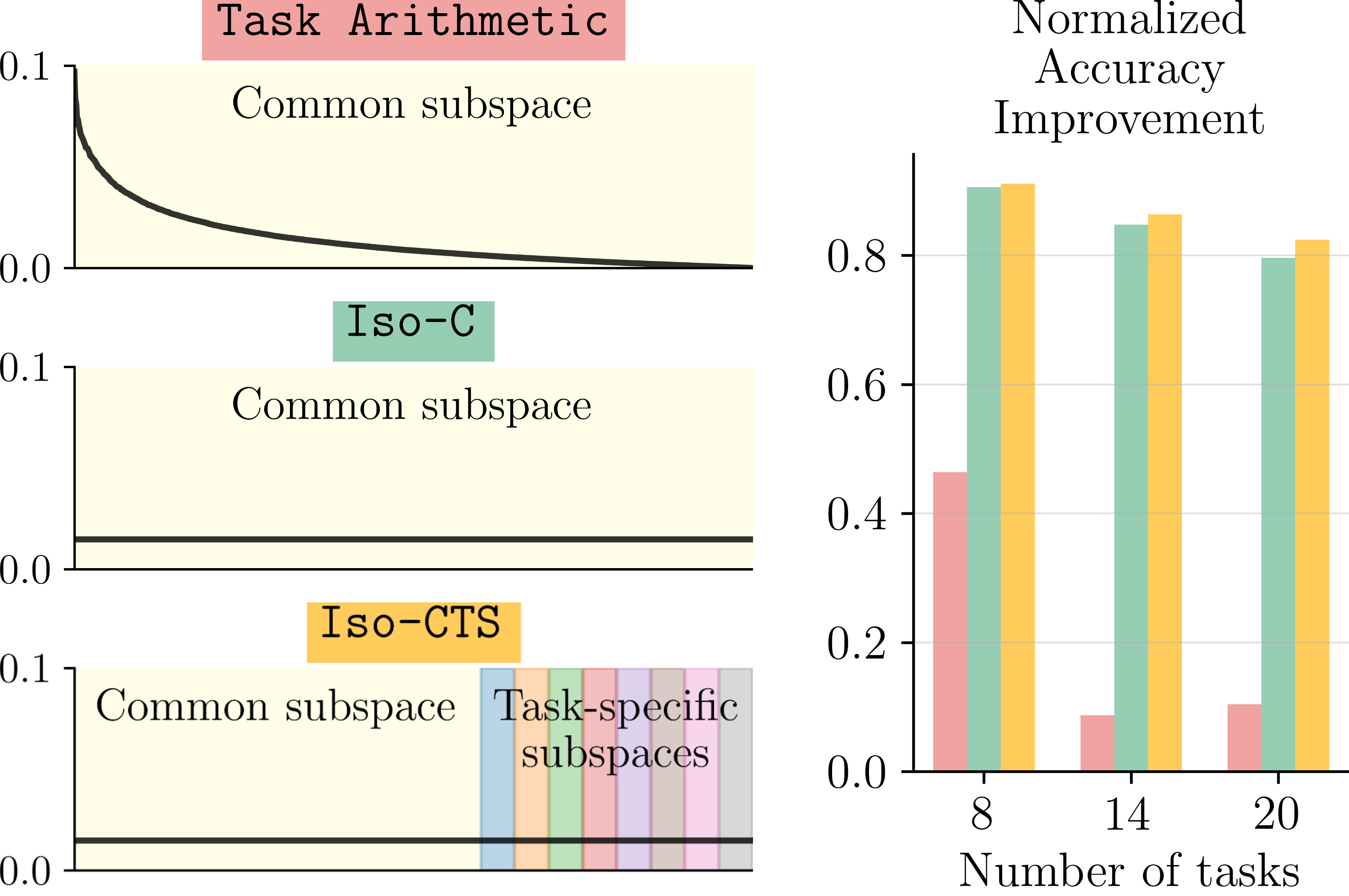}
    \caption{Spectrum of singular values for a single layer weight update matrix obtained by merging using \textcolor{ta}{\textbf{\texttt{Task Arithmetic}}} (top) compared to our approaches: \textcolor{isoc}{\textbf{\texttt{Iso-C}}} (middle) and \textcolor{isocts}{\textbf{\texttt{Iso-CTS}}} (bottom). \textcolor{ta}{\textbf{\texttt{Task Arithmetic}}} sums the task-specific matrices, which result in a spectrum with a few dominant components. \textcolor{isoc}{\textbf{\texttt{Iso-C}}} instead replaces this spectrum with a uniform one, which results in significant performance improvement. \textcolor{isocts}{\textbf{\texttt{Iso-CTS}}} enhances the common subspace with task-specific subspaces and yields state-of-the-art model merging performance.}
    \label{fig:teaser}
\end{figure}

Pre-trained models are the foundation of modern machine learning systems~\cite{carion2020endtoend, radford2021learning, DINO, zhai2023sigmoid}. In practice, they are typically fine-tuned for specialization on specific tasks~\cite{wortsman2022robust, ilharco2022patching}.
Recently, a growing body of research has focused on \textit{model merging}~\cite{li2023deep}, which combines multiple task-specific experts into a single multi-task model. Many methods have been proposed to improve the effectiveness of model merging by reducing sign conflicts~\cite{yadav2023tiesmerging}, by aligning gradients~\cite{DaheimMPGK24}, or through magnitude-based selection~\cite{MarczakTTC24}.
However, a significant performance gap between the combined and single-task models remains.

A key insight from \citet{ilharco2023task} is that \textit{task vectors}, defined as the offset between the \textit{flattened} fine-tuned weights and the pre-trained checkpoint, from different tasks are typically close to orthogonal. This orthogonality has been seen as a fundamental property enabling effective merging with reduced interference and has inspired works that enforce the orthogonality by modifying the fine-tuning procedure~\cite{PoYAW24}. Most recently, \citet{stoica2024knots} and \citet{tsv} have shown that accounting for the structure of the weight update matrix, dubbed \textit{task matrix}, is a more effective strategy for improving the performance of model merging. In this paper, we investigate precisely what the characteristics of task matrices are that favor effective model merging. Different from previous works, we propose to analyze the alignment between task-specific and merged subspaces.

Specifically, to capture the similarity between task matrices, we propose to investigate the \textit{Subspace Alignment Ratio}. Through the lens of Singular Value Decomposition, our metric quantifies the similarity between subspaces spanned by the top singular vectors of task matrices. When applied to compare matrices of the merged model to the task-specific ones, this metric strongly correlates with the performance of the merged model on a given task. This allows us to identify the directions amplified by multiple tasks as well as the underrepresented directions that lead to poor performance on corresponding tasks.

Our goal is to design a model merging technique that balances directions in the weight space across different tasks. We achieve this by flattening the singular values spectrum of the merged matrix, making it more uniform. Enforcing a uniform (isotropic) spectrum significantly improves the alignment and performance of the merged model. This simple yet effective adjustment, which requires no changes to the fine-tuning procedure, leads to substantial gains in merging performance (see method \texttt{Iso-C} in \cref{fig:teaser}).

However, tasks with dominant directions of smaller intensity compared to the majority of tasks and whose directions are orthogonal to the common directions may still remain underrepresented, especially when the number of tasks increases. To address this, we enhance isotropic model merging by introducing task-specific subspaces that retain unique task features while preserving shared knowledge. Our approach begins with the top singular values of the common subspace and iteratively replaces the least significant singular vectors with task-specific directions. This strategy allows us to increase the scalability of our merging approach to more tasks (see method \texttt{Iso-CTS} in \cref{fig:teaser}).

The main contributions of this paper are:
\begin{itemize}
\item We show that the alignment between the subspace spanned by the principal directions of the task-specific matrices and that of the merged matrix positively correlates with the performance of the merged model.
\item We demonstrate that applying an isotropic scaling to singular directions of merged task matrices improves the alignment between merged and task-specific matrices. This results in a simple yet highly effective technique for model merging that we call \texttt{Iso-C}, which outperforms most baselines.
\item We further enhance our approach by incorporating task-specific directions into the merged matrix resulting in \texttt{Iso-CTS}, a merging method that achieves state-of-the-art results, in particular for a large number of tasks.
\item Our methods demonstrate versatility, achieving state-of-the-art on vision and language merging benchmarks for both fully and LoRA fine-tuned models\footnote{The code is available at~\url{https://github.com/danielm1405/iso-merging}.}.
\end{itemize}

\section{Related Work}

\minisection{Model merging.}
Pre-trained models serve as a foundation for expert models specialized in specific downstream tasks~\cite{radford2021learning}. Recently, model merging has emerged as a promising technique to combine multiple expert models into a single multi-task model.
One of the pioneering works in the field, Task Arithmetic (TA)~\cite{ilharco2023task}, proposed to compute a \textit{task vector} as a difference between the expert and the pre-trained model and to then aggregate task vectors via scaled addition to create an expert in multiple tasks. The significant performance gap between individual experts and the combined model sparked an abundance of works with the aim of reducing interference when merging models. TIES~\cite{yadav2023tiesmerging} proposed a novel way to reduce sign conflicts between the parameters of expert models, Model Breadcrumbs~\cite{davari2023model} removed outliers from the task vectors, and Consensus Merging~\cite{wang2024localizing} removed catastrophic and selfish weights. These methods focused on per-parameter techniques to mitigate the interference, treating each parameter independently.

The aforementioned \textit{static merging} methods output a single set of multi-task weights which can be used as a drop-in replacement for the pre-trained model. However, a number of recent methods, dubbed \textit{dynamic merging}, alter the inference procedure to improve the results. Twin-Merging~\cite{lu2024twin0merging0} composes task-specific components at test-time and alters the inference algorithm requiring two forward passes. EMR-Merging~\cite{huang2024emr0merging0} uses additional per-task parameter masks and rescalers to perform inference. In this paper, we consider static merging exclusively.

\minisection{Singular Value Decomposition of model weights.}
While SVD of weight matrices has been primarily used for model compression~\cite{NIPS2014_2afe4567,KimPYCYS15}, recently its effectiveness was also identified for fine-tuning of large models. LoRA~\cite{hu2021lora} uses SVD to identify the similarities of weight updates between low-rank and full-rank fine-tuning.
MiLORA~\cite{milora} identifies that the bottom singular components correspond to noisy or long-tail information, while the top singular vectors contain important knowledge. Therefore, they propose a fine-tuning approach that updates only the minor singular components of the weight matrix while keeping the top singular components frozen. 
SVFT~\cite{svft} computes outer products of its singular vectors and, during fine-tuning updates, only sparse coefficients of these combinations.

\minisection{SVD for model merging.}
The structure imposed by SVD was used for model merging in KnOTS~\cite{stoica2024knots}, which proposes to concatenate the task-specific low-rank adaptation matrices (LoRA) and average the right-singular vectors before SVD reconstruction to obtain the merged weights.
The most similar work to us is the parallel work Task Singular Vectors (TSV)~\cite{tsv}, which measures task interference based on the interaction of singular vectors from different tasks and uses it to increase merging effectiveness. We share the motivation to improve model merging through SVD decomposition. However, while they focus on the orthogonalization of task-specific subspaces to reduce interference, we show that making singular values uniform in a common subspace is a surprisingly powerful method. Further, we show how to combine shared and task-specific subspaces for improved performance.

\section{Background and Motivation}
In this Section, we first describe the general framework of model merging and provide the notation used throughout the rest of the paper. We then motivate our approach via an analysis of the correlation between task similarity and performance improvement of the merged model.

\subsection{Model Merging}
Model merging integrates multiple deep neural network models, each individually trained (i.e. fine-tuned) on distinct tasks starting from the same pre-trained model, into a single merged model. Let $\theta_0$ denote the weights of the pre-trained network, and $\theta_t$ denote the fine-tuned weights for task $t$, with $t=1,\ldots, T$, where $T$ is the total number of tasks. We will use the notation $\theta_t^{(l)}$ to identify the weights of layer $l$ for task $t$ and $L$ to denote the total number of layers in a network. The objective of model merging is to find a merging function $f$, such that the model:
\begin{equation}
    \theta_{\text{M}}^{(\ell)} = f(\theta_0^{(\ell)},\{\theta_t^{(\ell)}\}_{t=1}^T), \quad \forall \ell=1,\ldots, L
\end{equation}
is able to perform all tasks on which the individual models $\theta_t$ are trained. 

Building upon Task Arithmetic (TA), we define the layer-wise \textit{task matrix} $\Delta^{(\ell)}_t$ as the difference between the weights of the model $\theta_t$ and the pre-trained model $\theta_0$ for layer $\ell$:
\begin{equation}
\Delta^{(\ell)}_t = \theta^{(\ell)}_{t} - \theta^{(\ell)}_{0}.
\end{equation}
In the rest of the paper, the $\ell$ superscript is omitted when not relevant to the discussion, and all definitions refer to an arbitrary layer. The authors of Task Arithmetic propose to solve the problem of model merging by defining a merging function that sums all task matrices to the pre-trained model weights:
\begin{equation}
\label{eq:ta}
    \theta^{(\ell)}_{\text{TA}} = \theta^{(\ell)}_0 + \alpha \Delta^{(\ell)}_{\text{TA}},
\end{equation}
where $\alpha$ is a scaling factor determined on a held-out validation dataset and $\Delta^{(\ell)}_{\text{TA}} = \sum_{t=1}^T \Delta^{(\ell)}_t$. The advantage of this merging strategy is that it allows for the reuse and transfer of knowledge from many fine-tuned models to the pre-trained model without requiring additional training or access to the original training data~\cite{ilharco2023task}.

\subsection{Cosine Similarity and Performance Improvement are Uncorrelated}

\begin{figure}[t!]
    \centering
    \subfloat[Cosine similarity between pairs of task vectors.]{\includegraphics[width=0.49\linewidth]{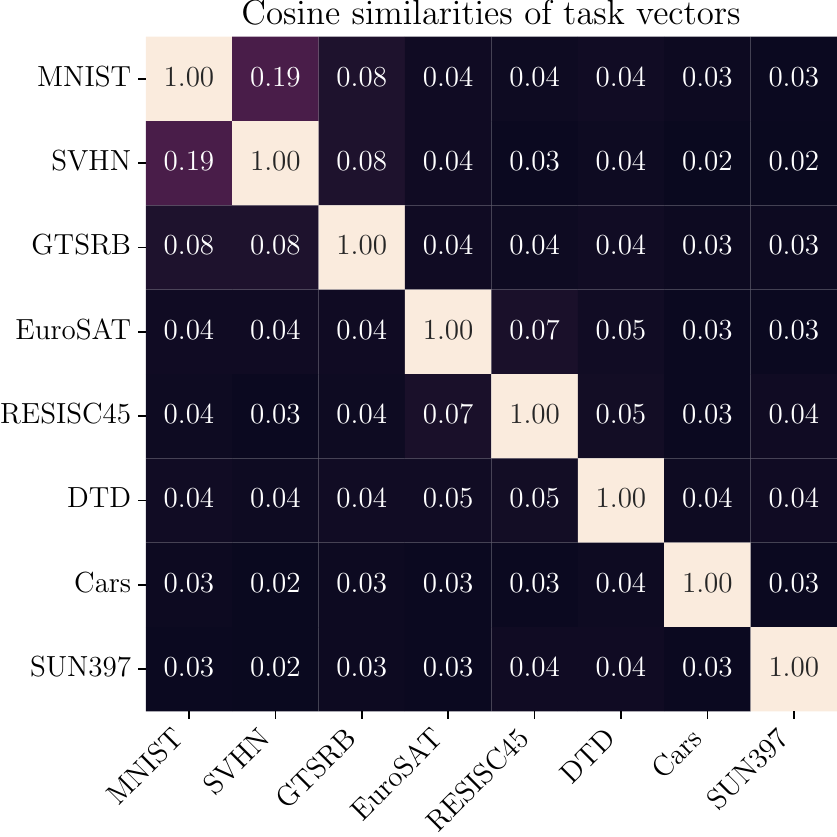}\label{fig:task-cossim}}
    \hfill
    \subfloat[NAI vs cosine similarity between task and merged vectors.]{\includegraphics[width=0.49\linewidth]{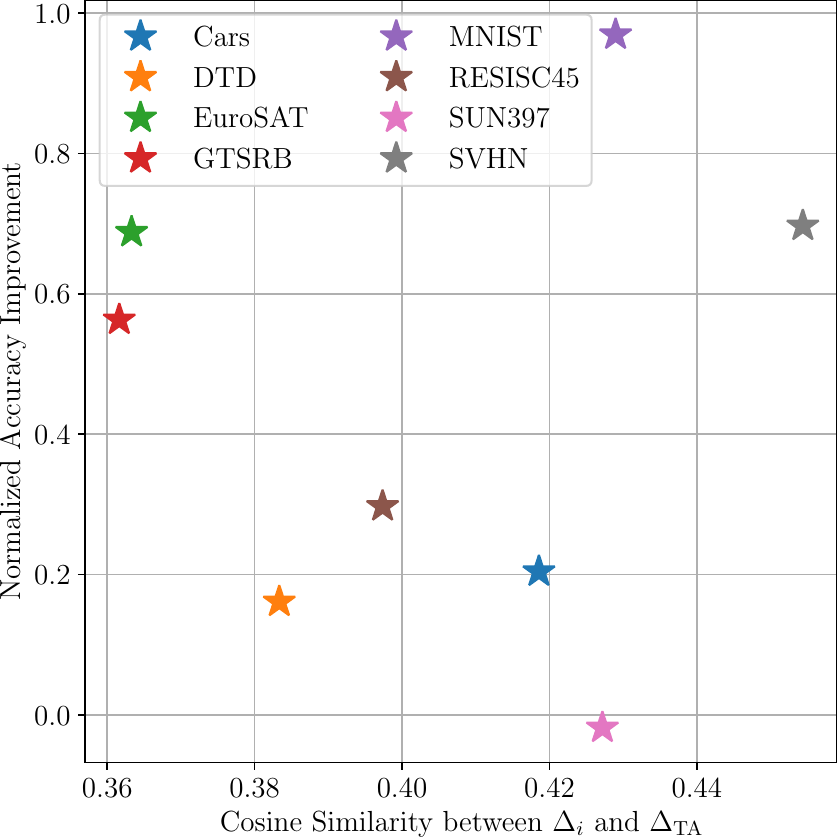}\label{fig:nai_vs_cossim_ta}}
    \caption{(a) Tasks vectors are typically close to \textit{orthogonal} to each other. (b) Models with very different normalized accuracy improvements (NAI) exhibit very close cosine similarities, and the correlation between cosine similarity and NAI is low.}
    \label{fig:cos_sim}
\end{figure}

Starting from the definition of Task Arithmetic (TA) in Eq.~\eqref{eq:ta}, we aim to explore the possible reasons for the improvement achieved by TA merging over the pre-trained (or zero-shot) model across multiple tasks. To empirically quantify performance gain, we propose the \textit{Normalized Accuracy Improvement (NAI)} metric, defined as: 
\begin{equation}
\label{eq:nai}
   \text{NAI}( \theta_{\text{M}}, \theta_t; \theta_0) = \frac{\text{Acc}(\theta_{\text{M}}) - \text{Acc}(\theta_0)}{\text{Acc}(\theta_{t}) - \text{Acc}(\theta_0)},
\end{equation}
which quantifies the improvement of the merged model $\theta_{\text{M}}$ relative to that achieved by the task-specific model $\theta_t$, both measured with respect to the zero-shot baseline $\theta_0$.\footnote{NAI differs from Normalized Accuracy~\cite{ortizjimenez2023tangent} which does not account for zero-shot performance.}

\citet{ilharco2023task} hypothesize that minimal inter-task interference -- captured by near-zero cosine similarity between the vectorized representation of the task matrices, i.e., $\langle \text{vec}(\Delta_i), \text{vec}(\Delta_j) \rangle \approx 0$ for $i \neq j$ (see \cref{fig:cos_sim}a) -- explains the effectiveness of Task Arithmetic. To investigate this further, we examine whether the cosine similarity between each task vector and the merged Task Arithmetic vector, $\langle \text{vec}(\Delta_{\text{TA}}), \text{vec}(\Delta_{t}) \rangle$, serves as an indicator of performance improvement, as quantified by $\text{NAI}(\theta_{\text{TA}}, \theta_t; \theta_0)$. However, we observe no clear correlation (see \cref{fig:cos_sim}b), suggesting that cosine similarity alone does not fully explain the observed performance gains. This indicates that the improvement achieved by Task Arithmetic likely originates from other factors, which we unveil below through spectral analysis of the Task Arithmetic and task-specific matrices.

\subsection{Performance Correlates with Subspace Alignment} \label{sec:motivation-alignment}

\begin{figure*}[t!]
    \centering
    \subfloat[Normalized Accuracy Improvement (NAI) vs. Average Subspace Alignment Ratio ($\text{SAR}_\text{avg}$).]{\includegraphics[width=0.5\linewidth]{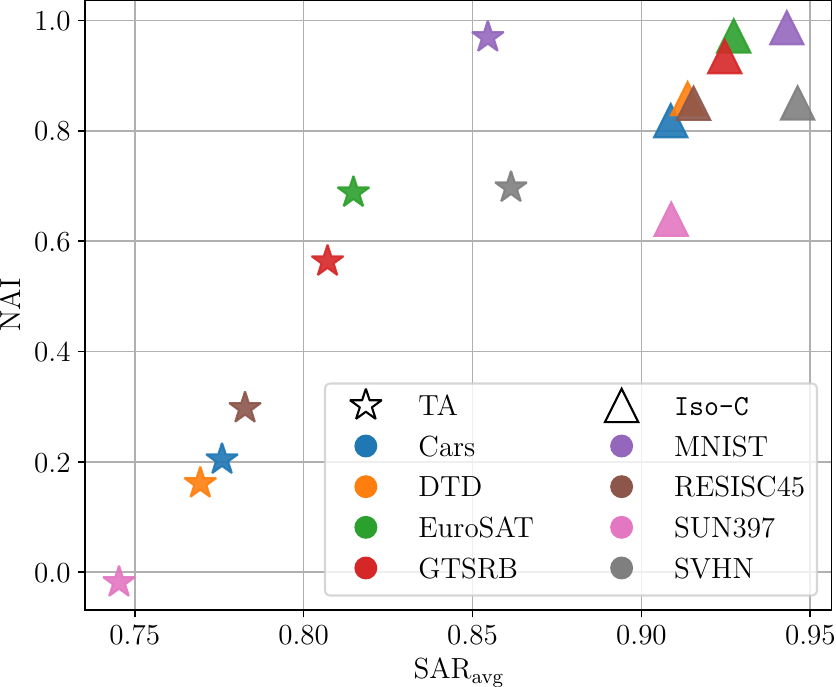}\label{fig:motivation_nai_vs_ar}}
    \hfill
    \subfloat[Average Subspace Alignment Ratios ($\text{SAR}_\text{avg}$) between pairs of task matrices.]{\includegraphics[width=0.43\linewidth]{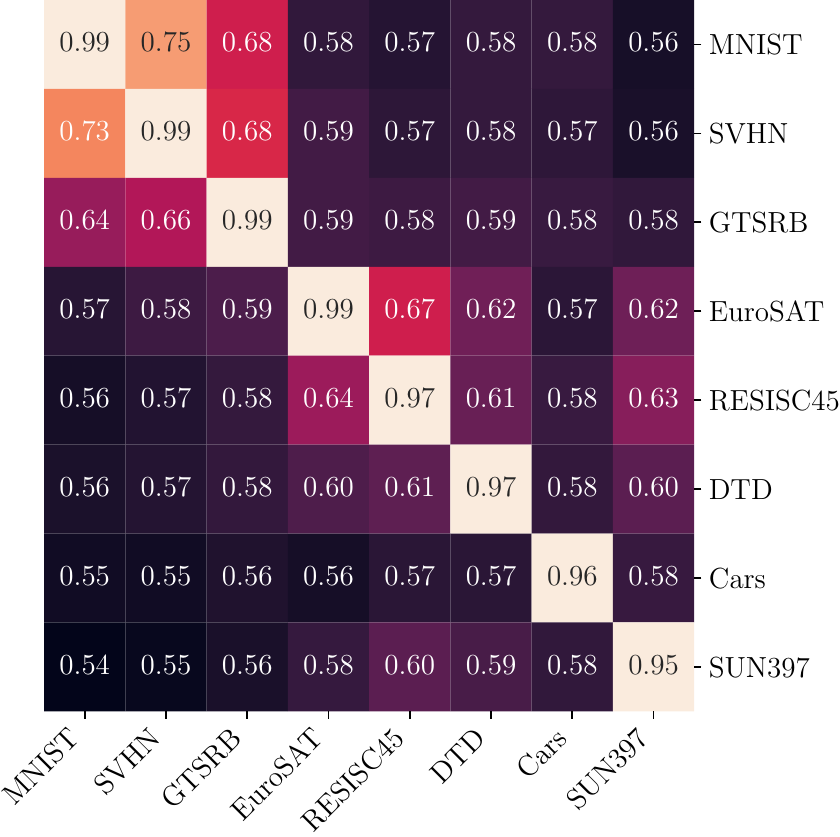}\label{fig:task-wise-alignment}}
    \caption{
    (a) NAI strongly correlates with $\text{SAR}_\text{avg}$ (Pearson correlation coefficient $\rho_{\mathrm{TA}} = 0.94$).
    (b) Note the groups of highly aligned tasks such as \{MNIST, SVHN, GTSRB\} and \{EuroSAT, RESISC45\}. By comparing (b) and (a), the mutually aligned datasets exhibit higher alignment with the merged model and consequently achieve good performance. On the other hand, tasks with low mutual alignment, such as DTD, Cars, and SUN397, are less aligned with the merged model and achieve poor performance.
    }
    \label{fig:nai_and_alignment}
\end{figure*}

We argue that the improvement in Task Arithmetic performance derives from the relationship between the top singular vectors of $\Delta_{\text{TA}}$ and those of each $\Delta_t$. Specifically, we hypothesize that the subspace of $\Delta_{\text{TA}}$ approximates the union of the subspaces of each $\Delta_t$, and that the overlap of this overall subspace with each task matrix correlates with the performance improvement of the merged model.

In order to empirically quantify the overlap between subspaces, we propose the \textit{Subspace Alignment Ratio (SAR)} metric.
We define SAR between a task matrix $\Delta_{t}$ and a generic merged task matrix $\Delta_{\text{M}}$ as:
\begin{equation}
\label{eq:sar}
    \text{SAR} (\Delta_{t}, \Delta_{\text{M}}; k_{\text{M}}) = \frac{||\Pi_{k_{\text{M}}, \text{M}}\Delta_{t}||_{F}}{||\Delta_{t} ||_{F}},
\end{equation}
where $\Pi_{k_{\text{M}},\text{M}} = U_{k_{\text{M}},\text{M}} U^\top_{k_{\text{M}},\text{M}}$ is the projection matrix onto the subspace spanned by the top $k_{\text{M}}$ left-singular vectors of $\Delta_{\text{M}}$. The columns of $U_{k_{\text{M}}, \text{M}}$ are obtained from the SVD decomposition of $\Delta_{\text{M}}$, and the number of singular vectors used ($k_{\text{M}}$) is determined from the merged task matrix $\Delta_{\text{M}}$ by minimizing the approximation error with $\epsilon=0.05$:
\begin{align}
\label{eq:rank}
  k_{M} &= \min \left\lbrace k : \left\| \Delta_{\text{M}} - \Pi_{k, \text{M}} \Delta_{\text{M}} \right\|_F \leq \epsilon \left\| \Delta_{\text{M}} \right\|_F \right\rbrace \nonumber \\
        &= \min \left\lbrace k : \frac{\sum_{i=k+1}^r \sigma_i^2}{\sum_{i=1}^r \sigma_i^2} \leq \epsilon^2 \right\rbrace,
\end{align}
where $\Sigma = \text{diag}(\sigma_1, \ldots, \sigma_r)$ contains the singular values of $\Delta_{\text{M}}$, and the equivalence follows from the definition of the Frobenius norm (see Appendix \ref{apdx:skewness-kM-SAR}).

SAR quantifies the alignment between the subspaces of two task matrices as a function of the number of dominant singular vectors of the merged matrix. To provide a single score measuring the overlap between two models, we denote with $\text{SAR}_\text{avg}$ the \textit{Average Subspace Alignment Ratio} across all layers.

In \cref{fig:motivation_nai_vs_ar} (left, represented by stars), we plot the Normalized Accuracy Improvement achieved by TA on each task, given by $\text{NAI}(\theta_{\text{TA}}, \theta_t; \theta_0)$, against the Average Subspace Alignment Ratio of each task matrix $\Delta_t$ with the merged task matrix $\Delta_{\text{TA}}$, i.e. $\text{SAR}_{\text{avg}}(\Delta_t, \Delta_\text{TA}; k_\text{TA})$. First, we note that the alignment between task and merged matrices are notably high (ranging from 0.75 to 0.87), but vary significantly across datasets. This suggests that task vectors are well represented in the subspace identified by the task-arithmetic matrix but with different degrees of alignment and consistency depending on dataset characteristics. Furthermore, we highlight a strong correlation (Pearson correlation coefficient $\rho_{\mathrm{TA}} = 0.94$) between the performance improvement on individual tasks achieved by $\theta_{\text{TA}}$ and the degree of alignment of $\Delta_t$ with $\Delta_{\text{TA}}$.

Analogous to the pairwise cosine similarity analysis between task vectors performed by \citet{ilharco2023task},  in~\cref{fig:task-wise-alignment} we measure the SAR between pairs of task matrices, $\text{SAR}{\text{avg}}(\Delta_i, \Delta_j; k_{\text{TA}})$, using the  $k_{\text{TA}}$ dominant components of the merged Task Arithmetic model.
Some groups of tasks exhibit higher alignment which is due to their semantic similarity, e.g. MNIST, SVHN, and GTSRB are digit recognition datasets, while EuroSAT and RESISC45 are satellite image datasets. On the other hand, datasets such as Cars, DTD or SUN397 are less aligned to other tasks. Most importantly, tasks belonging to highly aligned groups are also highly aligned with the TA model and achieve the highest accuracy improvements (see \cref{fig:motivation_nai_vs_ar}). The tasks that are not aligned are underrepresented in the dominant subspace of $\Delta_{\text{TA}}$, and the performance on them is low.

Based on the observed correlation between performance and alignment ratio, we hypothesize that a merging method that aims to achieve high alignment will also achieve strong performance. Therefore, in the next Section, we propose an approach called \textit{Isotropic Merging} that improves alignment and, most importantly, the performance of the merged models.

\section{Isotropic Merging in Common and Task-specific Subspaces}

In this Section, we propose a novel model merging method we call Isotropic Merging in Common and Task-Specific Subspaces (\texttt{Iso-CTS}). First, we introduce Isotropic Merging in Common Subspace (\texttt{Iso-C}), which is able to enhance the normalized accuracy improvement and the alignment of each task matrix using common directions identified by Task Arithmetic. Then, we show how to further enhance the performance of merged models by introducing task-specific directions to improve merging performance on sets of many diverse tasks.

\subsection{Isotropic Merging in Common Subspace}
\label{sec:isoc}

In~\cref{sec:motivation-alignment}, we demonstrated the high alignment of each task matrix with the matrix obtained by Task Arithmetic. This alignment indicates that the span of dominant singular vectors of the merged matrix effectively covers the subspace of each task and provides a good approximation of the \textit{common subspace}.
However, significant variability in the average alignment ratio across the dataset leads to a lower accuracy improvement for less aligned tasks compared to the tasks belonging to groups with high alignment. This variability stems from the skewness of the task arithmetic spectrum (\cref{fig:teaser} and \ref{fig:combined-spectra}), which is concentrated in the first few singular values (which we call \textit{top} or \textit{dominant}), favoring the tasks from the highly aligned groups.
Our proposed methodology, which we call \textit{\underline{Iso}tropic Merging in \underline{C}ommon Subspace} (\texttt{Iso-C}), aims to equalize the spectrum of the task arithmetic matrix in order to enhance the \textit{average subspace alignment ratio} and ensure a more balanced representation across tasks in the merged model. 

\begin{algorithm}[t]
    \caption{\texttt{Iso-C}: Isotropic Merging in Common Subspace}
    \label{alg:method_common}
    \begin{algorithmic}[1]
        \REQUIRE Task matrices $\Delta_1, \dots, \Delta_T \,\, \text{with}\,\, \Delta_t \in \mathbb{R}^{m \times n}$
        \STATE Sum task matrices: $\Delta_{\text{TA}} = \sum_{t=1}^{T} \Delta_{t}$
        \STATE Compute the SVD of $\Delta_{\text{TA}}$: $\Delta_{\text{TA}} = U \Sigma V^\top$, with
        $U \in \mathbb{R}^{m\times r}, \Sigma \in \mathbb{R}^{r \times r}, V \in \mathbb{R}^{n \times r}, \sigma = \text{diag}(\Sigma)\!\!\in \mathbb{R}^r$
        \STATE  Calculate isotropic factor:  \! $\overline{\sigma}=\frac{1}{r} \sum_{i=1}^{r} \sigma_i$ \hfill (Eq.\ref{eq:iso-c})
        \STATE Reconstruct the matrix: $\Delta_{\texttt{Iso-C}} = \overline{\sigma} U V^\top$ \hfill (Eq.\ref{eq:final-isoc})
        \STATE \textbf{return} $\Delta_{\texttt{Iso-C}}$
    \end{algorithmic}
\end{algorithm}
 
Consider the sum of task matrices $\Delta_{\text{TA}} = \sum_t \Delta_t$, where $\Delta_t \in \mathbb{R}^{m \times n}$. Via Singular Value Decomposition (SVD) on $\Delta_{\text{TA}}$ we obtain $\Delta_{\text{TA}} = U \Sigma V^{\top}$, where $U \in \mathbb{R}^{m \times r}$ and $V \in \mathbb{R}^{n \times r}$ represent, respectively, the left and right singular vectors of $\Delta_{\text{TA}}$, and $\Sigma \in \mathbb{R}^{r \times r}$ is the diagonal matrix containing the singular values. We denote the vector of singular values by $\sigma = \text{diag}(\Sigma) \in \mathbb{R}^{r}$. 

To reduce the skewness towards the dominant singular vectors of $\Delta_{\text{TA}}$,
we propose scaling all directions of the transformation applied by the right-singular vectors $V$ to a fixed value rather than using their corresponding singular values.  This ensures that the final transformation is \textit{isotropic}, with the scaling factor set to the average singular value:
\begin{equation}
\label{eq:iso-c}
\overline{\sigma} = \frac{1}{r}\sum_{i=1}^{r} \sigma_i,
\end{equation}
and merged matrix is computed using the reconstruction:
\begin{equation}
\label{eq:final-isoc}
    \Delta_{\texttt{Iso-C}} = \overline{\sigma} U V^\top.
\end{equation}
We apply this operation to all network layers, and the final merged model is defined as:
\begin{equation}
    \label{eq:isoc-alpha}
    \theta_{\texttt{Iso-C}}^{(\ell)} = \theta_0^{(\ell)} + \alpha \Delta^{(\ell)}_{\texttt{Iso-C}}, \,\,\, \forall \ell=1,\ldots, L
\end{equation}
where $\alpha$ is chosen on a held-out validation set.

Applying isotropic merging results in an enhancement of the normalized accuracy improvement and subspace alignment ratio (SAR) compared to Task Arithmetic (see Figure~\ref{fig:motivation_nai_vs_ar}). The increase in SAR is due to a higher number of dominant components $k_{\texttt{Iso-c}}$ in $\Delta_{\texttt{Iso-c}}$ (see~\cref{eq:rank}), derived from the singular vectors of $\Delta_{\text{TA}}$, which are aligned with the subspaces of individual tasks (see \Cref{apdx:iso-sar} for details). In \Cref{apdx:sar-interference}, we show that increased SAR is associated with reduced inter-task interference, measured by changes in internal activations induced by merging. In \cref{alg:method_common}, we present the \texttt{Iso-C} algorithm for a single layer.

\subsection{Isotropic Merging in Common and Task-Specific Subspaces}
\label{sec:isocts}

\begin{algorithm}[t]
    \caption{\texttt{Iso-CTS}: Isotropic Merging in Common and Task-Specific Subspaces (\textcolor{isoc}{green} -- shared with \texttt{Iso-C})}
    \label{alg:method_common_and_task_specific}
    \begin{algorithmic}[1]
        \REQUIRE Task matrices $\Delta_1, \dots, \Delta_T \,\, \text{with}\,\, \Delta_t \in \mathbb{R}^{m \times n}$
        \textcolor{isoc}{\STATE Sum task matrices $\Delta_{\text{TA}} = \sum_{t=1}^{T} \Delta_{t}$}
        \textcolor{isoc}{\STATE Compute the SVD of  $\Delta_{\text{TA}}$:
           $\Delta_{\text{TA}} = U \Sigma V^\top$, with} 
           \textcolor{isoc}{$U \in \mathbb{R}^{m\times r}, \Sigma \in \mathbb{R}^{r \times r}, V \in \mathbb{R}^{n \times r},\sigma = \text{diag}(\Sigma)\!\!\in \mathbb{R}^r$}
        \STATE \textbf{Retain} top-$k$ singular vectors and values from \textbf{common subspace}:
     $U^{1:k} = [u_1| \ldots | u_k] \quad V^{1:k} = [v_1| \ldots | v_k]$ 
     \quad \quad \quad \quad \quad $\sigma^{\text{cm}}=\text{diag}(\Sigma)^{1:k}$
        \STATE \textbf{Accumulate} task-specific directions via projection:
        \FOR{$t = 1$ to $T$}
        \STATE  $\overline{\Delta}_t = \Delta_t - U^{1:k} (U^{1:k})^{\top} \Delta_t$ \hfill (Eq.\ref{eq:residual})
        \STATE Compute SVD: $ \overline{\Delta}_{t} = \overline{U}_{t} \overline{\Sigma}_{t} \overline{V}^\top_{t}$
        \STATE Retain first $s=\frac{r-k}{T}$ components of $\overline{U}_{t}$ and $\overline{V}_{t}$:
        $\overline{U}_t^{1:s} = [\overline{u}_{t,1}\vert \ldots \vert \overline{u}_{t, s}] \quad \overline{V}_t^{1:s} = [\overline{v}_{t,1}\vert \ldots \vert \overline{v}_{t, s}]$
         \quad \quad \quad \quad \quad $\sigma_t^{\text{ts}}=\text{diag}(\overline{\Sigma}_t)^{1:s}$
        \ENDFOR
        \STATE \textbf{Combine} common and task-specific spaces: \vspace{-0.9em}\begin{align*}
          U_* &= [U^{1:k} \vert \overline{U}_1^{1:s}\vert  \ldots  \vert \overline{U}_T^{1:s}] \in \mathbb{R}^{m \times r}  \\[-3pt]  V_* &= [V^{1:k} \vert \overline{V}_1^{1:s}\vert\ldots\vert \overline{V}_T^{1:s}] \in \mathbb{R}^{n \times r} 
  \end{align*}
  \vspace{-2.0em}
       \STATE \textbf{Orthogonalize} $U_{*}$ and $V_{*}$ via whitening \hfill (Eq.\ref{eq:withening})
        \textcolor{isoc}{\STATE Calculate isotropic factor 
        $\overline{\sigma}$:}
        \vspace{-0.8em}
        \begin{equation*}
          \overline{\sigma} = \frac{1}{r}\Big(\sum_{i=1}^k \sigma_i^\text{cm} + \sum_{t=1}^{T} \sum_{i=1}^{s} \sigma_{t,i}^{\text{ts}}\Big) \quad \quad \quad \quad \quad \quad \text{(Eq.\ref{eq:iso-cts})}
         \end{equation*} 
         \vspace{-1.2em}
        \textcolor{isoc}{\STATE \textbf{Reconstruct}  the matrix
        $\Delta_{\texttt{Iso-CTS}} = \overline{\sigma} U_{*} V_{*}^\top$} \hfill (Eq.\ref{eq:final-cts})
        \STATE \textbf{return} $\Delta_{\texttt{Iso-CTS}}$
    \end{algorithmic}
\end{algorithm}

\begin{table*}[t!]
  \caption{\texttt{Iso-CTS} achieves state-of-the-art performance for all backbones on all evaluated scenarios. We present average absolute accuracy and average normalized accuracy (in subscript) in $\%$. The best method in \textbf{bold} and the second-best \underline{underlined}.}
  \label{tab:multitask_acc}
  \vspace{0.03in}
  \resizebox{1\textwidth}{!}{
    \begin{tabular}{cccc|ccc|ccc}
      \toprule
      \multirow{2}{*}{\textbf{Method}}         & \multicolumn{3}{c}{ViT-B/32}                     & \multicolumn{3}{c}{ViT-B/16}                     & \multicolumn{3}{c}{ViT-L/14}                    \\ 
                                               \cmidrule{2-10}
                                               & 8 tasks           & 14 tasks          & 20 tasks           & 8 tasks           & 14 tasks          & 20 tasks          & 8 tasks           & 14 tasks          & 20 tasks          \\ 
                                               \midrule
      \rowcolor{gray!15}
      \multicolumn{1}{c|}{Zero-shot} & $48.3$ & $57.2$ & $56.1$ & $55.3$ & $61.3$ & $59.7$ & $64.7$ & $68.2$ & $65.2$ \\
      \rowcolor{gray!15}
      \multicolumn{1}{c|}{Fine-tuned} & $92.8$ & $90.9$ & $91.3$ & $94.6$ & $92.8$ & $93.2$ & $95.8$ & $94.3$ & $94.7$ \\
      \midrule
      \multicolumn{1}{c|}{Weight Averaging}    & $66.3_{(72.1)}$ & $64.3_{(71.1)}$ & $61.0_{(67.5)}$ & $72.2_{(76.6)}$ & $69.5_{(74.8)}$ & $65.3_{(70.4)}$ & $79.6_{(83.2)}$ & $76.7_{(81.1)}$ & $71.6_{(75.6)}$ \\
      \multicolumn{1}{c|}{Task Arithmetic}     & $70.8_{(76.5)}$ & $65.3_{(72.1)}$ & $60.5_{(66.8)}$ & $75.4_{(79.6)}$ & $70.5_{(75.9)}$ & $65.8_{(70.8)}$ & $84.9_{(88.7)}$ & $79.4_{(84.0)}$ & $74.0_{(78.1)}$ \\
      \multicolumn{1}{c|}{TIES} & $75.1_{(81.0)}$ & $68.0_{(74.8)}$ & $63.4_{(69.9)}$ & $79.7_{(84.3)}$ & $73.2_{(78.7)}$ & $68.2_{(73.3)}$ & $86.9_{(90.7)}$ & $79.5_{(84.1)}$ & $75.7_{(79.8)}$ \\
      \multicolumn{1}{c|}{Consensus TA}        & $75.0_{(80.8)}$ & $70.4_{(77.4)}$ & $65.4_{(72.0)}$ & $79.4_{(83.9)}$ & $74.4_{(79.9)}$ & $69.8_{(74.9)}$ & $86.3_{(90.1)}$ & $82.2_{(86.9)}$ & $79.0_{(83.2)}$ \\
      \multicolumn{1}{c|}{TSV-M} & $85.9_{(92.3)}$ & $80.1_{(87.9)}$ & $\underline{77.1_{(84.3)}}$ & $89.0_{(93.9)}$ & $84.6_{(91.0)}$ & $\underline{80.6_{(86.5)}}$ & $93.0_{(97.0)}$ & $89.2_{(94.4)}$ & $\underline{87.7_{(92.5)}}$ \\
      \rowcolor{isoc!20}
      \multicolumn{1}{c|}{\textbf{\texttt{Iso-C} (Ours)}} & $\mathbf{86.3_{(92.9)}}$ & $\underline{80.3_{(88.1)}}$ & $75.5_{(82.5)}$ & $\underline{90.6_{(95.6)}}$ & $\underline{84.8_{(91.1)}}$ & $79.6_{(85.4)}$ & $\underline{94.2_{(98.3)}}$ & $\underline{89.3_{(94.5)}}$ & $87.6_{(92.2)}$ \\
      \rowcolor{isocts!25}
      \multicolumn{1}{c|}{\textbf{\texttt{Iso-CTS} (Ours)}} & $\underline{86.2_{(92.8)}}$ & $\mathbf{81.7_{(89.7)}}$ & $\mathbf{78.1_{(85.5)}}$ & $\mathbf{91.1_{(96.1)}}$ & $\mathbf{86.4_{(92.8)}}$ & $\mathbf{82.4_{(88.4)}}$ & $\mathbf{94.7_{(98.8)}}$ & $\mathbf{91.0_{(96.3)}}$ & $\mathbf{90.1_{(94.9)}}$ \\
      \bottomrule
    \end{tabular}
  }
  \vspace{-0.15in}
\end{table*}

The effectiveness of \texttt{Iso-C} depends on how well the common subspace -- identified by the dominant singular vectors of $\Delta_{\text{TA}}$ -- approximates the subspaces of the individual tasks. The approximation error arises from how these tasks interact when summed. The top singular directions of $\Delta_{\text{TA}}$ capture only the dominant common variations, while singular vectors associated with near-zero singular values provide negligible information. At the same time, tasks with dominant directions of smaller intensity compared to the majority of tasks and whose directions are orthogonal to the common directions remain underrepresented. This limitation becomes more pronounced as the number of tasks increases and the tasks become more diverse (see \Cref{apdx:iso-c-limitations} for an extended discussion).

To address this limitation, we propose enhancing the range of directions used by \texttt{Iso-C} to ensure that the task-specific directions, which are orthogonal to those of the common subspace, are incorporated into the singular basis of the final merged matrix. We call this methodology as \textit{\underline{Iso}tropic Merging in \underline{C}ommon and \underline{T}ask-\underline{S}pecific Subspaces} (\texttt{Iso-CTS}). 

Our approach starts with the top singular values of the common subspace and iteratively replaces the singular vectors associated with the lowest singular values with task-specific directions. The final goal is to find two orthonormal matrices $U_{*} \in \mathbb{R}^{m \times r}$ and $V_{*} \in \mathbb{R}^{n \times r}$ whose columns contain both common and task-specific directions. Afterward, the final matrix is reconstructed, and isotropic merging is applied. In the following, we provide a detailed explanation of our proposed algorithm.

\minisection{Retaining components from the common subspace.}  We retain the top-$k$ singular vectors associated with the subspace identified by $\Delta_{\text{TA}}$:
\begin{equation*}
    U^{1:k} = [u_1| \ldots | u_k] \quad V^{1:k} = [v_1| \ldots | v_k],
\end{equation*}
where $U^{1:k}$, $V^{1:k}$ are the top-$k$ left- and right-singular vectors from the SVD of $\Delta_{\text{TA}}$. We analyze the impact of selecting $k$ in~\cref{sec:analysis}.

\minisection{Accumulating task-specific directions.} We project each task-specific matrix $\Delta_t$ onto the subspace orthogonal to the common subspace, i.e. the space spanned by top left-singular directions of the common subspace $U^{1:k}$:
 \begin{equation}
 \label{eq:residual}
     \overline{\Delta}_t = \Delta_t - U^{1:k} (U^{1:k})^T \Delta_t.
 \end{equation}
We then compute the SVD of $\overline{\Delta}_t = \overline{U}_t\, \overline{\Sigma}_t \overline{V}_t$ and retain the top $s=\frac{r-k}{T}$ directions for each task $t$:
\begin{equation*}
    \overline{U}_t^{1:s}\!\!=\![\overline{u}_{t,1}\vert \ldots \vert \overline{u}_{t,s}] \,\,\, \overline{V}_t^{1:s}\!\!=\![\overline{v}_{t,1}\vert \ldots \vert \overline{v}_{t,s}], \forall t=1,\!\ldots\!, T.
\end{equation*}
The orthogonal projection Eq.~\eqref{eq:residual} guarantees that both the left- and right-singular vectors of $\overline{\Delta}_t$, representing task-specific directions, are orthogonal to the subspace spanned by the common directions (given by $U^{1:k}$).

\minisection{Combining common and task-specific matrices.} After identifying the $k$ principal vectors for the common subspace and $s=\frac{r-k}{T}$ principal vectors for each task, we now combine the common and task-specific directions by concatenating them: $U_* = [U^{1:k}\vert\overline{U}_1^{1:s}\vert\ldots  \vert \overline{U}_T^{1:s}] \in \mathbb{R}^{m \times r} $ and $V_* = [V^{1:k} \vert \overline{V}_1^{1:s}\vert\ldots\vert\overline{V}_T^{1:s}] \in \mathbb{R}^{n \times r}$.

\minisection{Orthogonalization.} There is no guarantee that the left- and right-singular task-specific vectors are orthogonal to each other, as we are only projecting each task matrix onto the common subspace. To reconstruct the final merged matrix, we must orthogonalize $U_*$ and $V_*$. Following~\citet{tsv}, we compute the SVD of $U_*= P_{U_*} \Sigma_{U_*} Q_{U_*}^\top $ and $V_*=P_{V_*}  \Sigma_{V_*} Q_{V_*}^\top$, and whiten~\citep{schonemann1966}:
\begin{equation}
\label{eq:withening}
    U_* =  P_{U_*} Q_{U_*}^\top  \quad V_* = P_{V_*} Q_{V_*}^\top.
\end{equation}

\minisection{Isotropic scaling and reconstruction.} Finally, we reconstruct the final merged matrix and apply isotropic merging:
\begin{equation}   
\label{eq:final-cts}
\Delta_{\texttt{Iso-CTS}}=\overline{\sigma} U_{*} V_{*}^\top,
\end{equation}
where $\overline{\sigma}$ is obtained by averaging the singular values associated with the vectors selected for both common and task-specific subspaces. Specifically, defining $\sigma^{\text{cm}}= \text{diag}(\Sigma)^{1:k} \in \mathbb{R}^{k}$, the vector of singular values associated with the common subspace identified by $U_{1:k}$ and $V_{1:k}$, and $\sigma^{\text{ts}}_{t} = \text{diag}(\overline{\Sigma}_t)^{1:s}\in \mathbb{R}^{s}$, with $s=\frac{r-k}{T}$, the vector of singular values associated with each task-specific subspace $\overline{U}_t^{1:s}$ and $\overline{V}_t^{1:s}$, we define the scaling factor as:
\begin{equation}
\label{eq:iso-cts}
       \overline{\sigma} = \frac{1}{r}\Big(\sum_{i=1}^k \sigma_i^\text{cm} + \sum_{t=1}^{T} \sum_{i=1}^{s} \sigma_{t,i}^{\text{ts}}\Big).
\end{equation}
Finally, similar to \texttt{ISO-C}, the merged model is defined as:
\begin{equation}
    \label{eq:isocts-alpha}
    \theta_{\texttt{Iso-{CTS}}}^{(\ell)} = \theta_0^{(\ell)} + \alpha \Delta^{(\ell)}_{\texttt{Iso-{CTS}}}, \quad \forall \ell=1,\ldots,L
\end{equation}
where $\alpha$ is chosen on a held-out validation set.

\section{Experimental Results}
\label{sec:exp}

\begin{figure*}[t!]
    \vspace{-0.15in}
    \centering
    \subfloat[Spectra of singular values for different values of interpolation coefficient ($\beta$).]{\includegraphics[width=0.325\linewidth]{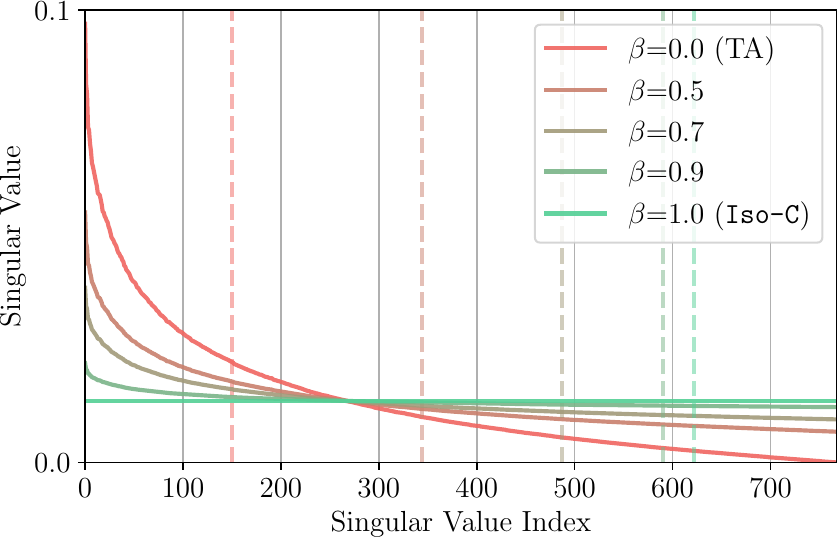}\label{fig:ta-iso-interpolation-spectrum}}
    \hfill
    \subfloat[Average Subspace Alignment Ratio ($\text{SAR}_{\text{avg}}$) vs. interpolation coefficient ($\beta$).]{\includegraphics[width=0.32\linewidth]{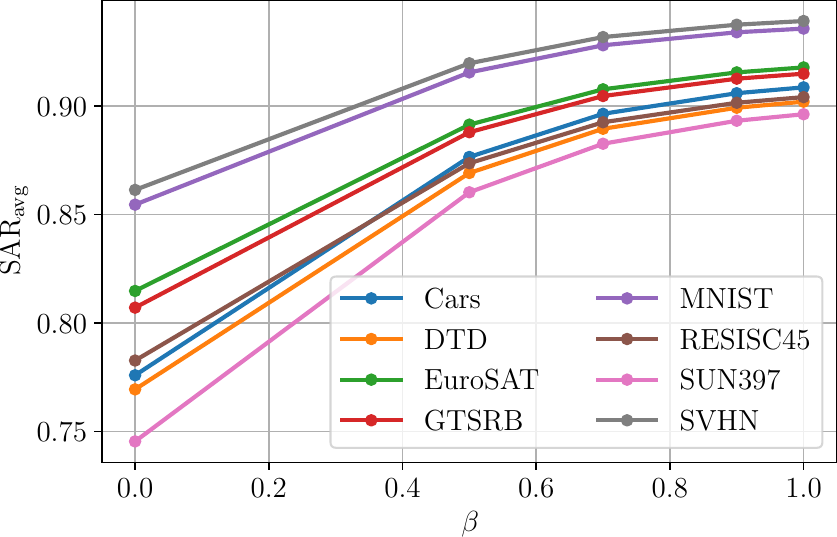}\label{fig:ta-iso-interpolation-ar-vs-beta}}
    \hfill
    \subfloat[Normalized Accuracy Improvement (NAI) vs. interpolation coefficient ($\beta$).]{\includegraphics[width=0.32\linewidth]{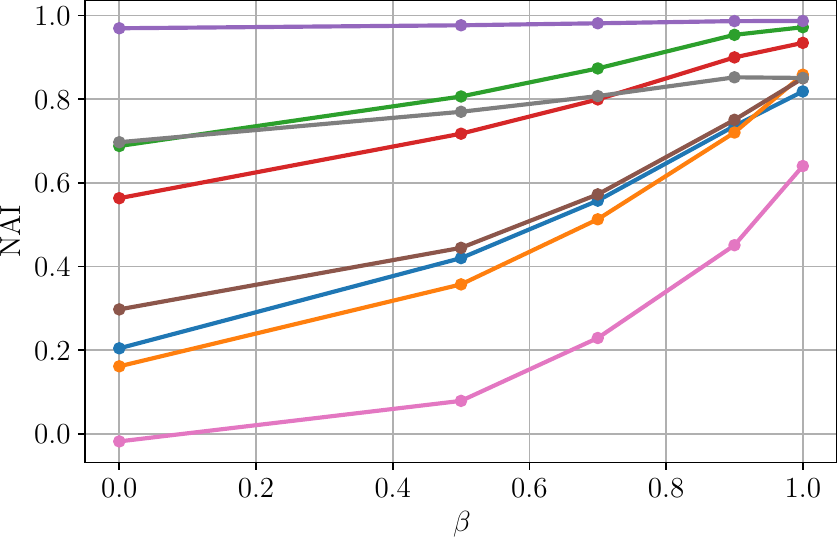}\label{fig:ta-iso-interpolation-nai-vs-beta}}
    \vspace{-0.05in}
    \caption{
    (a) Interpolating from $\Delta_{\text{TA}}$ ($\beta = 0$) towards $\Delta_{\texttt{Iso-C}}$ ($\beta = 1$) makes the spectrum of singular values of $\Delta_{\text{M}}$ more uniform and increases the number of preserved components $k_{\text{M}}$ (Eq.~\eqref{eq:rank}) denoted by dashed lines. (b) This results in an increased alignment between each task-specific model and merged model measured by $\text{SAR}_{\text{avg}}$. (c) As alignment increases, the performance also improves as predicted based on the strong correlation between these two properties investigated in~\cref{sec:motivation-alignment}.
    }
    \vspace{-0.25in}
    \label{fig:ta-iso-interpolation}
\end{figure*}

\subsection{Fully fine-tuned vision models}

We evaluate our approaches over sets of 8, 14, and 20 datasets, following~\citet{wang2024localizing}. We provide the details of the datasets in \cref{apdx:datasets}. We consider three variants of CLIP~\cite{radford2021learning} with ViT-B/32, ViT-B/16 and ViT-L/14 as visual encoders~\cite{dosovitskiy2021an}. We use the checkpoints fine-tuned on the tasks above, provided in~\citet{wang2024localizing} (see \Cref{apdx:additional-baselines} for results using TA checkpoints). If not stated otherwise, we present the results using the ViT-B/16 visual encoder.

We compare our approaches with the following model merging methods: weight averaging~\cite{wortsman2022model}, Task Arithmetic~\cite{ilharco2023task}, TIES-Merging~\cite{yadav2023tiesmerging}, Consensus TA~\cite{wang2024localizing} and TSV-M~\cite{tsv}. We include the results of the zero-shot model and fine-tuned models serving as lower- and upper-bound, respectively. We compare the results based on absolute and normalized accuracy following standard practice~\cite{wang2024localizing, tsv}.

\cref{tab:multitask_acc} presents our main results for multi-task model merging. \texttt{Iso-CTS} achieves state-of-the-art results in all of the settings. 
\texttt{Iso-C} achieves very similar results to \texttt{Iso-CTS} in the 8 task scenario. However, \texttt{Iso-CTS} significantly outperforms \texttt{Iso-C} when merging 14 and 20 models, with improvements of up to 2.8\% in absolute accuracy. This suggests that it is possible to faithfully represent a small number of tasks in the common subspace. However, when the number of tasks increases, it becomes crucial to retain important directions from the task-specific subspaces in order to maximize model merging effectiveness.

\subsection{LoRA-adapted vision models}

To evaluate our approaches in low-rank adaptation scenario, we follow the evaluation protocol of KnOTS~\cite{stoica2024knots}, a recent state-of-the-art method for merging LoRA fine-tuned models. We use codebase and checkpoints provided by KnOTS: ViT-B/32 and ViT-L/14 fine-tuned with rank 16 LoRA~\cite{hu2021lora} on 8 vision tasks. To adapt our methodologies to low-rank regime, we simply operate on reconstructed task matrices, i.e. $\Delta W_t = B_t A_t$, where $A_t, B_t$ are LoRA matrices for task $t$. We compare \texttt{Iso-C} and \texttt{Iso-CTS} with TIES and DARE-TIES~\cite{yu2024language} -- combined with KnOTS or not -- and TA.

We present the results in \Cref{tab:lora}. Our methods, which are general purpose merging techniques, significantly outperform KnOTS, which are specifically designed for the LoRA merging. This highlights the versatility of \texttt{Iso} methods.

\begin{table}[t]
    \centering
    \caption{Normalized per-task average accuracy. We merge 8 models fine-tuned with LoRA following~\cite{stoica2024knots}.}
    \label{tab:lora}
    \vspace{0.03in}
    \resizebox{0.35\textwidth}{!}{
    \begin{tabular}{l|cc}
    \toprule
    \textbf{Method} & ViT-B/32 & ViT-L/14 \\
    \midrule
    TA & 63.7 & 74.4 \\
    TIES & 63.7 & 75.2 \\
    DARE-TIES & 63.7 & 74.7 \\
    KnOTS-TIES & 68.0 & 78.2 \\
    KnOTS-DARE-TIES & 63.9 & 75.6 \\
    \midrule
    \rowcolor{isoc!20}
    \textbf{\texttt{Iso-C} (Ours)} & \underline{73.6} & \underline{83.7} \\
    \rowcolor{isocts!25}
    \textbf{\texttt{Iso-CTS} (Ours)} & \textbf{73.7} & \textbf{85.3} \\
    \bottomrule
    \end{tabular}}
    \vspace{-0.3in}
\end{table}

\subsection{Language models}

We present NLP results following the experimental setup from MaTS~\cite{tam2023merging}. We use T5-Large-LM-Adapt~\cite{lester2021power} base model (a variant of T5-Large~\cite{raffel2020exploring}) fine-tuned on subsets of 8 and 7 NLP tasks from T0 mixture~\cite{sanh2021multitask}. We compare our approaches with weight averaging, TA, TIES, Fisher Merging~\cite{matena2021merging}, RegMean~\cite{jin2023dataless}, and MaTS~\cite{tam2023merging}.

We present the results in \Cref{tab:nlp}. Both \texttt{Iso-C} and \texttt{Iso-CTS} significantly outperform the competing approaches, which highlights the versatility of our proposed methods. We observe that \texttt{Iso-CTS} achieves very similar results to \texttt{Iso-C} suggesting that the common space captures all the directions necessary to reliably represent these 7 and 8 NLP tasks.

\begin{table}[ht]
\centering
\caption{NLP results using T5-Large-LM-Adapt fine-tuned on tasks from T0 mixture. We present average absolute accuracy.}
\label{tab:nlp}
\vspace{0.03in}
\resizebox{0.41\textwidth}{!}{
\begin{tabular}{l|cc}
\toprule
\textbf{Method} 
  & \textbf{8 tasks} 
  & \textbf{7 tasks} \\
  & \footnotesize{\cite{zhou2022not}} & \footnotesize{\cite{yadav2023tiesmerging}} \\
\midrule
Fine-tuned                & 80.7 & 85.9 \\
\midrule
Weight Averaging          & 56.4 & 60.5 \\
Task Arithmetic           & 63.8 & 69.2 \\
TIES                      & 62.8 & 71.9 \\
Fisher Merging            & 57.7 & 61.0 \\
RegMean                   & 69.1 & 74.3 \\
MaTS                      & 72.5 & 81.5 \\
\midrule
\rowcolor{isoc!20}
\textbf{\texttt{Iso-C} (Ours)}              & \textbf{75.6} & \textbf{83.3} \\
\rowcolor{isocts!25}
\textbf{\texttt{Iso-CTS} (Ours)}            & \underline{75.2} & \underline{82.8} \\
\bottomrule
\end{tabular}}
\end{table}

\subsection{Analysis and Ablations}\label{sec:analysis}

All the experiments in this Section are conducted on fully fine-tuned ViT-B/16 models. In \Cref{apdx:computational-complexity} we provide the computational complexity analysis of our approaches.

\begin{figure*}[t!]
    \centering
    \subfloat[Normalized Accuracy Improvement (NAI) of a model created by retaining $k$ components of \texttt{Iso-C} (associated with top-$k$ singular vectors from $\Delta_{\text{TA}}$).]{\includegraphics[width=0.39\linewidth]{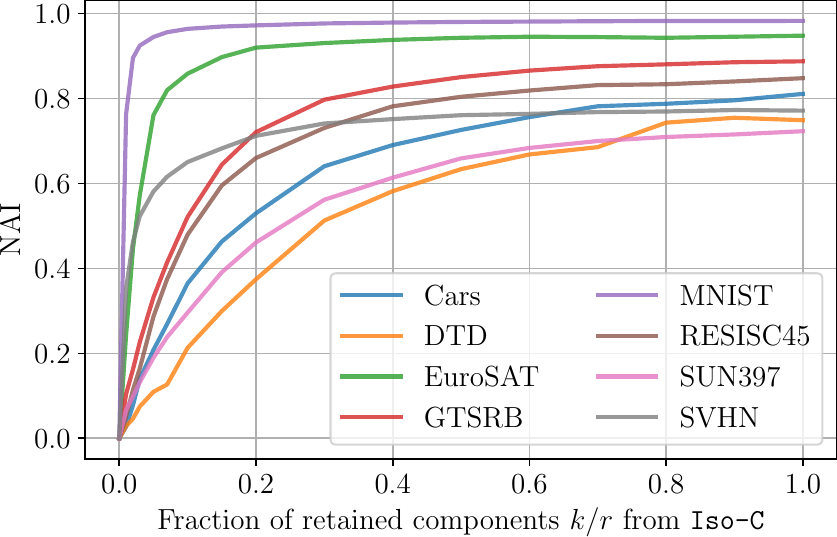}\label{fig:nai-vs-k}}
    \hfill
    \subfloat[Average Subspace Alignment Ratios ($\text{SAR}_{\text{avg}}$) between merged and task-specific models for varying sets of tasks.]{\includegraphics[width=0.289\linewidth]{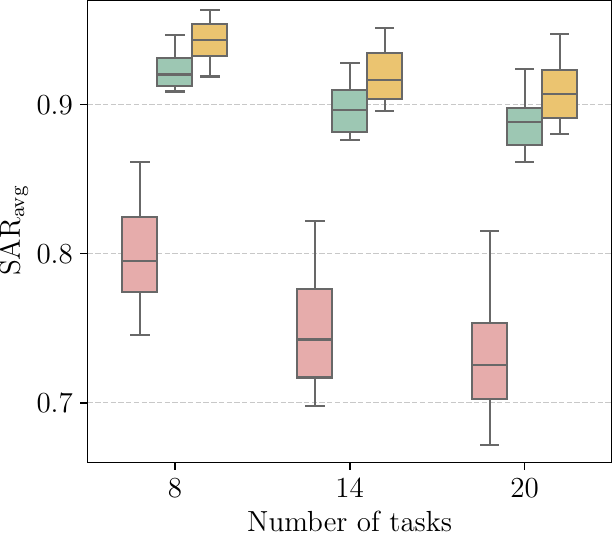}\label{fig:alignment-across-tasks}}
    \hfill
    \subfloat[Distribution of accuracies of the merged models for varying sets of tasks.]{\includegraphics[width=0.285\linewidth]{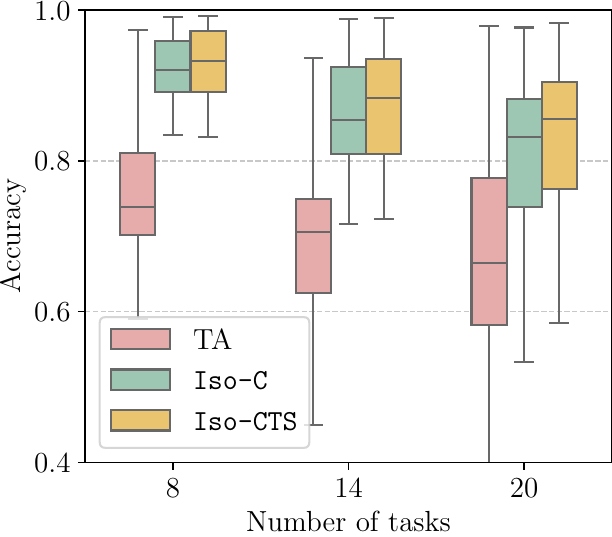}\label{fig:acc-across-tasks}}
    \vspace{-0.05in}
    \caption{
    (a) The directions associated with the least significant singular values of $\Delta_{\text{TA}}$ have a minor contribution to the performance of \texttt{Iso-C} model.
    (b) Task-specific directions introduced in \texttt{Iso-CTS} improve the Average Subspace Alignment Ratio ($\text{SAR}_{\text{avg}}$) between task-specific models and the merged model compared to \texttt{Iso-C} which uses only a common subspace.
    (c) Higher alignment translates to higher accuracy of \texttt{Iso-CTS} with respect to \texttt{Iso-C}.
    }
    \vspace{-0.2in}
    \label{fig:iso-cts-motivation}
\end{figure*}

\minisection{From Task Arithmetic to Isotropic Merging.}
We analyze what happens when interpolating between the singular values obtained by Task Arithmetic (TA) and those obtained by \texttt{Iso-C}, i.e. the model with the following spectra:
\begin{equation}
    \Sigma_{\beta} = (1 - \beta) \Sigma_{\text{TA}} + \beta \Sigma_{\texttt{Iso-C}},
\end{equation}
where $\beta$ is an interpolation coefficient. Firstly, \cref{fig:ta-iso-interpolation-spectrum} presents the change in singular values spectrum as we interpolate towards $\Delta_{\texttt{Iso-C}}$ ($\beta \rightarrow 1$). The skewed spectrum achieved by Task Arithmetic becomes isotropic, i.e. the scaling factor is equal along all of the singular directions. In~\cref{fig:ta-iso-interpolation-ar-vs-beta} we observe a steady increase in alignment between task-specific and merged models as measured by $\text{SAR}_{\text{avg}}$ (Eq.~\eqref{eq:sar}), and \cref{fig:ta-iso-interpolation-nai-vs-beta} shows that as alignment increases (with $\beta \rightarrow 1$), the performance of the merged model improves across all tasks. These results are consistent with our findings from~\cref{sec:motivation-alignment} that show a strong correlation between alignment and the performance of the final model.

\minisection{The impact of singular directions on performance.}
We analyze which singular directions contribute to the improvement of individual tasks. We truncate the flattened spectrum of \texttt{Iso-C}, keeping the $k$ directions associated with the leftmost singular values, i.e. $\sigma_i = \overline{\sigma}$ for $i \leq k$ and $\sigma_i = 0$ for $i > k$. Note that the leftmost $k$ directions are the ones associated with the highest singular values of $\Delta_{\text{TA}}$. We plot the task-wise Normalized Accuracy Improvement (NAI, Eq. \eqref{eq:nai}) for varying $k$ in~\cref{fig:nai-vs-k}. We observe that the first few directions are responsible for rapid improvement on several tasks. Notably, these tasks belong to the aligned groups identified in~\cref{sec:motivation-alignment} such as \{MNIST, SVHN, GTSRB\} and \{EuroSAT, RESISC45\}. Moreover, the directions associated with the least significant singular values of $\Delta_{\text{TA}}$ have a negligible contribution to the performance. This supports our intuition for replacing less significant common directions with task-specific components in \texttt{Iso-CTS} (see ~\cref{sec:isocts}). \cref{fig:alignment-across-tasks} shows that \texttt{Iso-CTS} achieves higher Average Subspace Alignment Ratio ($\text{SAR}_{\text{avg}}$, Eq.~\eqref{eq:sar}) than \texttt{Iso-C}.
Most importantly, \cref{fig:acc-across-tasks} shows that thanks to the addition of task-specific directions, \texttt{Iso-CTS} achieves better performance across tasks.

\minisection{Size of the common subspace for \texttt{Iso-CTS}.}
\begin{figure}[t!]
    \centering
    \vspace{0.05in}
    \includegraphics[width=0.8\linewidth]{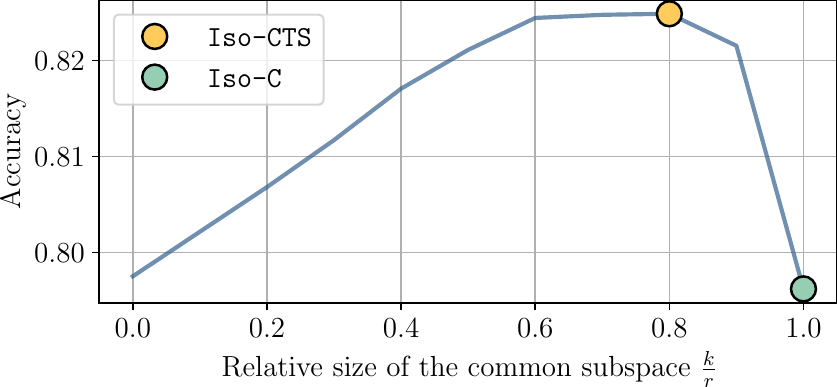}
    \vspace{-0.05in}
    \caption{\texttt{Iso-CTS} is robust to the selected size of the common subspace as any value leads to improvement over \texttt{Iso-C}. These results are for the 20-task scenario.}
    \label{fig:frac-common-subspace}
    \vspace{-0.1in}
\end{figure}
While \texttt{Iso-C} operates only in the common subspace, \texttt{Iso-CTS} enhances it with task-specific subspaces. Therefore, we must select the size of the common subspace $k$ (and consequently the size of each task-specific subspace given by $\frac{r-k}{T}$). \cref{fig:frac-common-subspace} plots the relationship between accuracy and the fraction of subspace assigned for the common subspace ($\frac{k}{r}$) when merging 20 tasks. When $\frac{k}{r}=1$ \texttt{Iso-CTS} is equivalent to \texttt{Iso-C} and suffers a 2.8\% drop in accuracy from the maximum. The optimal fraction of common subspace $\frac{k}{r} = 0.8$, and we use this as a default value for \texttt{Iso-CTS} across all settings. Moreover, note that \texttt{Iso-CTS} is quite robust to the selection of this hyperparameter -- any $\frac{k}{r} \in (0.0, 1.0)$ offers a performance improvement over \texttt{Iso-C} while the performance for $\frac{k}{r} \in [0.5, 0.9]$ varies by less than 0.5\% from the optimal one.

\section{Conclusion}

In this work, we introduced an isotropic model merging framework that enhances alignment between task-specific and merged model subspaces to significantly improve the multi-task performance of the final merged model. We proposed \texttt{Iso-C}, which leverages Singular Value Decomposition to equalize singular values and create a more balanced representation across tasks, and \texttt{Iso-CTS}, which further incorporates task-specific directions to retain unique task features while preserving shared knowledge. \texttt{Iso-CTS} achieves state-of-the-art results across multiple model scales and task sets, demonstrating that subspace alignment is a critical factor in effective model merging. These findings provide new insights into model merging and pave the way for the future development of more effective techniques to combine the knowledge of multiple models.

\minisection{Limitations.} The common subspace is determined by Task Arithmetic, which can be suboptimal, and better methods could be developed. Although the proposed methods achieve state-of-the-art results in the LoRA merging scenario, they could be adapted to leverage the low-rank structure of task matrices to further improve the performance and efficiency.

\section*{Acknowledgements}
Daniel Marczak is supported by National Centre of Science (NCN, Poland) Grant No. 2021/43/O/ST6/02482. This work was supported by Horizon Europe Programme under GA no. 101120237, project ``ELIAS: European Lighthouse of AI for Sustainability''. Simone Magistri acknowledges travel support from ELIAS (GA no 101120237). We acknowledge the Spanish project PID2022-143257NB-I00, financed by MCIN/AEI/10.13039/501100011033 and FEDER, and Funded by the European Union ELLIOT project. Bartłomiej Twardowski acknowledges the grant RYC2021-032765-I and National Centre of Science (NCN, Poland) Grant No. 2023/51/D/ST6/02846.
Andrew D. Bagdanov acknowledges funding support from the Italian national project ``Collaborative Explainable neuro-symbolic AI for Decision Support Assistant'', CAI4DSA, CUP B13C23005640006.

\section*{Impact Statement}
This paper aims to advance the field of Machine Learning, specifically the subfield focused on merging models fine-tuned on different tasks to create a more effective multi-task model. With the growing popularity of deep learning, increasingly powerful open-source models are becoming widely available and are being adopted in both research and industry. Advances in model merging could enhance the flexibility of utilizing these models by providing an efficient way to combine their specialized capabilities. Beyond this, our paper presents work whose goal is to advance the field of Machine Learning. There are many potential societal consequences of our work, none of which we feel must be specifically highlighted here.

\bibliography{main}

\begin{thebibliography}{59}
\providecommand{\natexlab}[1]{#1}
\providecommand{\url}[1]{\texttt{#1}}
\expandafter\ifx\csname urlstyle\endcsname\relax
  \providecommand{\doi}[1]{doi: #1}\else
  \providecommand{\doi}{doi: \begingroup \urlstyle{rm}\Url}\fi

\bibitem[Ba et~al.(2016)Ba, Kiros, and Hinton]{ba2016layer}
Ba, J.~L., Kiros, J.~R., and Hinton, G.~E.
\newblock Layer normalization.
\newblock \emph{arXiv preprint arXiv: 1607.06450}, 2016.

\bibitem[Bossard et~al.(2014)Bossard, Guillaumin, and Van~Gool]{bossard_food-101_2014}
Bossard, L., Guillaumin, M., and Van~Gool, L.
\newblock Food-101 – {Mining} {Discriminative} {Components} with {Random} {Forests}.
\newblock In \emph{ECCV}, 2014.

\bibitem[Carion et~al.(2020)Carion, Massa, Synnaeve, Usunier, Kirillov, and Zagoruyko]{carion2020endtoend}
Carion, N., Massa, F., Synnaeve, G., Usunier, N., Kirillov, A., and Zagoruyko, S.
\newblock End-to-end object detection with transformers.
\newblock \emph{ECCV}, 2020.

\bibitem[Caron et~al.(2021)Caron, Touvron, Misra, Jegou, Mairal, Bojanowski, and Joulin]{DINO}
Caron, M., Touvron, H., Misra, I., Jegou, H., Mairal, J., Bojanowski, P., and Joulin, A.
\newblock Emerging properties in self-supervised vision transformers.
\newblock \emph{ICCV}, 2021.

\bibitem[Cheng et~al.(2017)Cheng, Han, and Lu]{cheng2017remote}
Cheng, G., Han, J., and Lu, X.
\newblock Remote sensing image scene classification: Benchmark and state of the art.
\newblock \emph{Proceedings of the IEEE}, 2017.

\bibitem[Cimpoi et~al.(2014)Cimpoi, Maji, Kokkinos, Mohamed, and Vedaldi]{dtd}
Cimpoi, M., Maji, S., Kokkinos, I., Mohamed, S., and Vedaldi, A.
\newblock Describing textures in the wild.
\newblock In \emph{CVPR}, 2014.

\bibitem[Clanuwat et~al.(2018)Clanuwat, Bober-Irizar, Kitamoto, Lamb, Yamamoto, and Ha]{clanuwat_deep_2018}
Clanuwat, T., Bober-Irizar, M., Kitamoto, A., Lamb, A., Yamamoto, K., and Ha, D.
\newblock Deep {Learning} for {Classical} {Japanese} {Literature}.
\newblock \emph{arXiv preprint arXiv: 1607.06450}, 2018.

\bibitem[Coates et~al.(2011)Coates, Ng, and Lee]{coates_analysis_2011}
Coates, A., Ng, A., and Lee, H.
\newblock An {Analysis} of {Single}-{Layer} {Networks} in {Unsupervised} {Feature} {Learning}.
\newblock In \emph{Proceedings of the {Fourteenth} {International} {Conference} on {Artificial} {Intelligence} and {Statistics}}. JMLR Workshop and Conference Proceedings, 2011.

\bibitem[Cohen et~al.(2017)Cohen, Afshar, Tapson, and van Schaik]{cohen_emnist_2017}
Cohen, G., Afshar, S., Tapson, J., and van Schaik, A.
\newblock {EMNIST}: {Extending} {MNIST} to handwritten letters.
\newblock In \emph{IJCNN}, 2017.

\bibitem[Daheim et~al.(2024)Daheim, M{\"{o}}llenhoff, Ponti, Gurevych, and Khan]{DaheimMPGK24}
Daheim, N., M{\"{o}}llenhoff, T., Ponti, E.~M., Gurevych, I., and Khan, M.~E.
\newblock Model merging by uncertainty-based gradient matching.
\newblock In \emph{ICLR}, 2024.

\bibitem[Davari \& Belilovsky(2024)Davari and Belilovsky]{davari2023model}
Davari, M.-J. and Belilovsky, E.
\newblock Model breadcrumbs: Scaling multi-task model merging with sparse masks.
\newblock \emph{ECCV}, 2024.

\bibitem[Denton et~al.(2014)Denton, Zaremba, Bruna, LeCun, and Fergus]{NIPS2014_2afe4567}
Denton, E.~L., Zaremba, W., Bruna, J., LeCun, Y., and Fergus, R.
\newblock Exploiting linear structure within convolutional networks for efficient evaluation.
\newblock In \emph{NeurIPS}, 2014.

\bibitem[Dosovitskiy et~al.(2021)Dosovitskiy, Beyer, Kolesnikov, Weissenborn, Zhai, Unterthiner, Dehghani, Minderer, Heigold, Gelly, Uszkoreit, and Houlsby]{dosovitskiy2021an}
Dosovitskiy, A., Beyer, L., Kolesnikov, A., Weissenborn, D., Zhai, X., Unterthiner, T., Dehghani, M., Minderer, M., Heigold, G., Gelly, S., Uszkoreit, J., and Houlsby, N.
\newblock An image is worth 16x16 words: Transformers for image recognition at scale.
\newblock In \emph{ICLR}, 2021.

\bibitem[Du et~al.(2024)Du, Lee, Li, Jiang, Guo, Yu, Liu, Goh, Tang, He, and Zhang]{guodong24neurips}
Du, G., Lee, J., Li, J., Jiang, R., Guo, Y., Yu, S., Liu, H., Goh, S.~K., Tang, H.-K., He, D., and Zhang, M.
\newblock Parameter competition balancing for model merging.
\newblock In \emph{NeurIPS}, 2024.

\bibitem[Gargiulo et~al.(2025)Gargiulo, Crisostomi, Bucarelli, Scardapane, Silvestri, and Rodolà]{tsv}
Gargiulo, A.~A., Crisostomi, D., Bucarelli, M.~S., Scardapane, S., Silvestri, F., and Rodolà, E.
\newblock Task singular vectors: Reducing task interference in model merging.
\newblock In \emph{CVPR}, 2025.

\bibitem[Goodfellow et~al.(2013)Goodfellow, Erhan, Carrier, Courville, Mirza, Hamner, Cukierski, Tang, Thaler, Lee, Zhou, Ramaiah, Feng, Li, Wang, Athanasakis, Shawe-Taylor, Milakov, Park, Ionescu, Popescu, Grozea, Bergstra, Xie, Romaszko, Xu, Chuang, and Bengio]{goodfellow_challenges_2013}
Goodfellow, I.~J., Erhan, D., Carrier, P.~L., Courville, A., Mirza, M., Hamner, B., Cukierski, W., Tang, Y., Thaler, D., Lee, D.-H., Zhou, Y., Ramaiah, C., Feng, F., Li, R., Wang, X., Athanasakis, D., Shawe-Taylor, J., Milakov, M., Park, J., Ionescu, R., Popescu, M., Grozea, C., Bergstra, J., Xie, J., Romaszko, L., Xu, B., Chuang, Z., and Bengio, Y.
\newblock Challenges in {Representation} {Learning}: {A} {Report} on {Three} {Machine} {Learning} {Contests}.
\newblock \emph{Neural Networks}, 2013.

\bibitem[Helber et~al.(2019)Helber, Bischke, Dengel, and Borth]{eurosat}
Helber, P., Bischke, B., Dengel, A., and Borth, D.
\newblock Eurosat: A novel dataset and deep learning benchmark for land use and land cover classification.
\newblock \emph{Journal of Selected Topics in Applied Earth Observations and Remote Sensing}, 2019.

\bibitem[Hu et~al.(2021)Hu, Shen, Wallis, Allen-Zhu, Li, Wang, and Chen]{hu2021lora}
Hu, J.~E., Shen, Y., Wallis, P., Allen-Zhu, Z., Li, Y., Wang, S., and Chen, W.
\newblock Lora: Low-rank adaptation of large language models.
\newblock \emph{ICLR}, 2021.

\bibitem[Huang et~al.(2024)Huang, Ye, Chen, He, Yue, and Ouyang]{huang2024emr0merging0}
Huang, C., Ye, P., Chen, T., He, T., Yue, X., and Ouyang, W.
\newblock Emr-merging: Tuning-free high-performance model merging.
\newblock \emph{NeurIPS}, 2024.

\bibitem[Ilharco et~al.(2022)Ilharco, Wortsman, Gadre, Song, Hajishirzi, Kornblith, Farhadi, and Schmidt]{ilharco2022patching}
Ilharco, G., Wortsman, M., Gadre, S.~Y., Song, S., Hajishirzi, H., Kornblith, S., Farhadi, A., and Schmidt, L.
\newblock Patching open-vocabulary models by interpolating weights.
\newblock In \emph{NeurIPS}, 2022.

\bibitem[Ilharco et~al.(2023)Ilharco, Ribeiro, Wortsman, Schmidt, Hajishirzi, and Farhadi]{ilharco2023task}
Ilharco, G., Ribeiro, M.~T., Wortsman, M., Schmidt, L., Hajishirzi, H., and Farhadi, A.
\newblock Editing models with task arithmetic.
\newblock In \emph{ICLR}, 2023.

\bibitem[Jin et~al.(2023)Jin, Ren, Preotiuc{-}Pietro, and Cheng]{jin2023dataless}
Jin, X., Ren, X., Preotiuc{-}Pietro, D., and Cheng, P.
\newblock Dataless knowledge fusion by merging weights of language models.
\newblock In \emph{ICLR}, 2023.

\bibitem[Kim et~al.(2016)Kim, Park, Yoo, Choi, Yang, and Shin]{KimPYCYS15}
Kim, Y., Park, E., Yoo, S., Choi, T., Yang, L., and Shin, D.
\newblock Compression of deep convolutional neural networks for fast and low power mobile applications.
\newblock In \emph{ICLR}, 2016.

\bibitem[Krause et~al.(2013)Krause, Stark, Deng, and Fei-Fei]{cars}
Krause, J., Stark, M., Deng, J., and Fei-Fei, L.
\newblock {3D Object representations for fine-grained categorization}.
\newblock In \emph{ICCV Workshops}, 2013.

\bibitem[Krizhevsky \& Hinton(2009)Krizhevsky and Hinton]{krizhevsky_learning_nodate}
Krizhevsky, A. and Hinton, G.
\newblock Learning multiple layers of features from tiny images.
\newblock Technical Report~0, University of Toronto, Toronto, Ontario, 2009.
\newblock URL \url{https://www.cs.toronto.edu/~kriz/learning-features-2009-TR.pdf}.

\bibitem[Lecun et~al.(1998)Lecun, Bottou, Bengio, and Haffner]{MNIST}
Lecun, Y., Bottou, L., Bengio, Y., and Haffner, P.
\newblock Gradient-based learning applied to document recognition.
\newblock \emph{Proceedings of the IEEE}, 1998.

\bibitem[Lee et~al.(2025)Lee, Choi, Lee, Kim, and Hong]{lee2025adarankadaptiverankpruning}
Lee, C., Choi, J., Lee, C., Kim, D., and Hong, S.
\newblock Adarank: Adaptive rank pruning for enhanced model merging.
\newblock \emph{arXiv preprint arXiv: 2503.22178}, 2025.

\bibitem[Lester et~al.(2021)Lester, Al-Rfou, and Constant]{lester2021power}
Lester, B., Al-Rfou, R., and Constant, N.
\newblock The power of scale for parameter-efficient prompt tuning.
\newblock \emph{EMNLP}, 2021.

\bibitem[Li et~al.(2023)Li, Peng, Zhang, Ding, Hu, and Shen]{li2023deep}
Li, W., Peng, Y., Zhang, M., Ding, L., Hu, H., and Shen, L.
\newblock Deep model fusion: A survey.
\newblock \emph{arXiv preprint arXiv: 2309.15698}, 2023.

\bibitem[Lingam et~al.(2024)Lingam, Tejaswi, Vavre, Shetty, Gudur, Ghosh, Dimakis, Choi, Bojchevski, and Sanghavi]{svft}
Lingam, V., Tejaswi, A., Vavre, A., Shetty, A., Gudur, G.~K., Ghosh, J., Dimakis, A., Choi, E., Bojchevski, A., and Sanghavi, S.
\newblock {SVFT:} parameter-efficient fine-tuning with singular vectors.
\newblock \emph{CoRR}, abs/2405.19597, 2024.

\bibitem[Lu et~al.(2024)Lu, Fan, Wei, Qu, Chen, and Cheng]{lu2024twin0merging0}
Lu, Z., Fan, C., Wei, W., Qu, X., Chen, D., and Cheng, Y.
\newblock Twin-merging: Dynamic integration of modular expertise in model merging.
\newblock \emph{NeurIPS}, 2024.

\bibitem[Marczak et~al.(2024)Marczak, Twardowski, Trzcinski, and Cygert]{MarczakTTC24}
Marczak, D., Twardowski, B., Trzcinski, T., and Cygert, S.
\newblock {MagMax:} {Leveraging} {Model} {Merging} for {Seamless} {Continual} {Learning}.
\newblock In \emph{ECCV}, 2024.

\bibitem[Matena \& Raffel(2021)Matena and Raffel]{matena2021merging}
Matena, M. and Raffel, C.
\newblock Merging models with fisher-weighted averaging.
\newblock In \emph{NeurIPS}, 2021.

\bibitem[Netzer et~al.(2011)Netzer, Wang, Coates, Bissacco, Wu, and Ng]{svhn}
Netzer, Y., Wang, T., Coates, A., Bissacco, A., Wu, B., and Ng, A.~Y.
\newblock Reading digits in natural images with unsupervised feature learning.
\newblock In \emph{NeurIPS Workshops}, 2011.

\bibitem[Nilsback \& Zisserman(2008)Nilsback and Zisserman]{nilsback_automated_2008}
Nilsback, M.-E. and Zisserman, A.
\newblock Automated {Flower} {Classification} over a {Large} {Number} of {Classes}.
\newblock In \emph{2008 {Sixth} {Indian} {Conference} on {Computer} {Vision}, {Graphics} \& {Image} {Processing}}, 2008.

\bibitem[Ortiz{-}Jim{\'{e}}nez et~al.(2023)Ortiz{-}Jim{\'{e}}nez, Favero, and Frossard]{ortizjimenez2023tangent}
Ortiz{-}Jim{\'{e}}nez, G., Favero, A., and Frossard, P.
\newblock Task arithmetic in the tangent space: Improved editing of pre-trained models.
\newblock In \emph{NeurIPS}, 2023.

\bibitem[Parkhi et~al.(2012)Parkhi, Vedaldi, Zisserman, and Jawahar]{parkhi_cats_2012}
Parkhi, O.~M., Vedaldi, A., Zisserman, A., and Jawahar, C.~V.
\newblock Cats and dogs.
\newblock In \emph{CVPR}, 2012.

\bibitem[Po et~al.(2024)Po, Yang, Aberman, and Wetzstein]{PoYAW24}
Po, R., Yang, G., Aberman, K., and Wetzstein, G.
\newblock Orthogonal adaptation for modular customization of diffusion models.
\newblock In \emph{CVPR}, 2024.

\bibitem[Radford et~al.(2021)Radford, Kim, Hallacy, Ramesh, Goh, Agarwal, Sastry, Askell, Mishkin, Clark, Krueger, and Sutskever]{radford2021learning}
Radford, A., Kim, J.~W., Hallacy, C., Ramesh, A., Goh, G., Agarwal, S., Sastry, G., Askell, A., Mishkin, P., Clark, J., Krueger, G., and Sutskever, I.
\newblock Learning transferable visual models from natural language supervision.
\newblock In \emph{ICML}, 2021.

\bibitem[Raffel et~al.(2020)Raffel, Shazeer, Roberts, Lee, Narang, Matena, Zhou, Li, and Liu]{raffel2020exploring}
Raffel, C., Shazeer, N., Roberts, A., Lee, K., Narang, S., Matena, M., Zhou, Y., Li, W., and Liu, P.~J.
\newblock Exploring the limits of transfer learning with a unified text-to-text transformer.
\newblock \emph{Journal of machine learning research}, 2020.

\bibitem[Sanh et~al.(2022)Sanh, Webson, Raffel, Bach, Sutawika, Alyafeai, Chaffin, Stiegler, Scao, Raja, et~al.]{sanh2021multitask}
Sanh, V., Webson, A., Raffel, C., Bach, S.~H., Sutawika, L., Alyafeai, Z., Chaffin, A., Stiegler, A., Scao, T.~L., Raja, A., et~al.
\newblock Multitask prompted training enables zero-shot task generalization.
\newblock \emph{ICLR}, 2022.

\bibitem[Schönemann(1966)]{schonemann1966}
Schönemann, P.~H.
\newblock A generalized solution of the orthogonal procrustes problem.
\newblock \emph{Psychometrika}, 1966.

\bibitem[Socher et~al.(2013)Socher, Perelygin, Wu, Chuang, Manning, Ng, and Potts]{socher_recursive_nodate}
Socher, R., Perelygin, A., Wu, J., Chuang, J., Manning, C.~D., Ng, A., and Potts, C.
\newblock Recursive deep models for semantic compositionality over a sentiment treebank.
\newblock In \emph{EMNLP}, 2013.

\bibitem[Stallkamp et~al.(2011)Stallkamp, Schlipsing, Salmen, and Igel]{gtsrb}
Stallkamp, J., Schlipsing, M., Salmen, J., and Igel, C.
\newblock The german traffic sign recognition benchmark: a multi-class classification competition.
\newblock In \emph{IJCNN}, 2011.

\bibitem[Stoica et~al.(2025)Stoica, Ramesh, Ecsedi, Choshen, and Hoffman]{stoica2024knots}
Stoica, G., Ramesh, P., Ecsedi, B., Choshen, L., and Hoffman, J.
\newblock {Model merging with SVD to tie the Knots}.
\newblock In \emph{ICLR}, 2025.

\bibitem[Tam et~al.(2023)Tam, Bansal, and Raffel]{tam2023merging}
Tam, D., Bansal, M., and Raffel, C.
\newblock Merging by matching models in task subspaces.
\newblock \emph{TMLR}, 2023.

\bibitem[Vasudevan \& Ramakrishna(2017)Vasudevan and Ramakrishna]{vasudevan2017hierarchical}
Vasudevan, V. and Ramakrishna, M.
\newblock A hierarchical singular value decomposition algorithm for low rank matrices.
\newblock \emph{arXiv preprint arXiv: 1710.02812}, 2017.

\bibitem[Veeling et~al.(2018)Veeling, Linmans, Winkens, Cohen, and Welling]{veeling_rotation_2018}
Veeling, B.~S., Linmans, J., Winkens, J., Cohen, T., and Welling, M.
\newblock Rotation {Equivariant} {CNNs} for {Digital} {Pathology}.
\newblock In \emph{MICCAI}, 2018.

\bibitem[Wang et~al.(2024{\natexlab{a}})Wang, Xiao, Li, Wang, Chen, and Chen]{milora}
Wang, H., Xiao, Z., Li, Y., Wang, S., Chen, G., and Chen, Y.
\newblock Milora: Harnessing minor singular components for parameter-efficient {LLM} finetuning.
\newblock \emph{CoRR}, abs/2406.09044, 2024{\natexlab{a}}.

\bibitem[Wang et~al.(2024{\natexlab{b}})Wang, Dimitriadis, Ortiz{-}Jim{\'{e}}nez, Fleuret, and Frossard]{wang2024localizing}
Wang, K., Dimitriadis, N., Ortiz{-}Jim{\'{e}}nez, G., Fleuret, F., and Frossard, P.
\newblock Localizing task information for improved model merging and compression.
\newblock In \emph{ICML}, 2024{\natexlab{b}}.

\bibitem[Wortsman et~al.(2022{\natexlab{a}})Wortsman, Ilharco, Gadre, Roelofs, Gontijo-Lopes, Morcos, Namkoong, Farhadi, Carmon, Kornblith, et~al.]{wortsman2022model}
Wortsman, M., Ilharco, G., Gadre, S.~Y., Roelofs, R., Gontijo-Lopes, R., Morcos, A.~S., Namkoong, H., Farhadi, A., Carmon, Y., Kornblith, S., et~al.
\newblock Model soups: averaging weights of multiple fine-tuned models improves accuracy without increasing inference time.
\newblock In \emph{ICML}, 2022{\natexlab{a}}.

\bibitem[Wortsman et~al.(2022{\natexlab{b}})Wortsman, Ilharco, Kim, Li, Kornblith, Roelofs, Lopes, Hajishirzi, Farhadi, Namkoong, and Schmidt]{wortsman2022robust}
Wortsman, M., Ilharco, G., Kim, J.~W., Li, M., Kornblith, S., Roelofs, R., Lopes, R.~G., Hajishirzi, H., Farhadi, A., Namkoong, H., and Schmidt, L.
\newblock Robust fine-tuning of zero-shot models.
\newblock In \emph{CVPR}, 2022{\natexlab{b}}.

\bibitem[Xiao et~al.(2017)Xiao, Rasul, and Vollgraf]{xiao_fashion-mnist_2017}
Xiao, H., Rasul, K., and Vollgraf, R.
\newblock Fashion-mnist: a novel image dataset for benchmarking machine learning algorithms.
\newblock \emph{arXiv preprint arXiv: 1708.07747}, 2017.

\bibitem[Xiao et~al.(2016)Xiao, Ehinger, Hays, Torralba, and Oliva]{sun397}
Xiao, J., Ehinger, K.~A., Hays, J., Torralba, A., and Oliva, A.
\newblock Sun database: Exploring a large collection of scene categories.
\newblock \emph{IJCV}, 2016.

\bibitem[Yadav et~al.(2023)Yadav, Tam, Choshen, Raffel, and Bansal]{yadav2023tiesmerging}
Yadav, P., Tam, D., Choshen, L., Raffel, C., and Bansal, M.
\newblock {TIES}-merging: Resolving interference when merging models.
\newblock In \emph{NeurIPS}, 2023.

\bibitem[Yang et~al.(2024)Yang, Shen, Wang, Guo, Chen, Wang, and Tao]{yang2024representation}
Yang, E., Shen, L., Wang, Z., Guo, G., Chen, X., Wang, X., and Tao, D.
\newblock Representation surgery for multi-task model merging.
\newblock \emph{ICML}, 2024.

\bibitem[Yu et~al.(2024)Yu, Yu, Yu, Huang, and Li]{yu2024language}
Yu, L., Yu, B., Yu, H., Huang, F., and Li, Y.
\newblock Language models are super mario: Absorbing abilities from homologous models as a free lunch.
\newblock In \emph{ICML}, 2024.

\bibitem[Zhai et~al.(2023)Zhai, Mustafa, Kolesnikov, and Beyer]{zhai2023sigmoid}
Zhai, X., Mustafa, B., Kolesnikov, A., and Beyer, L.
\newblock Sigmoid loss for language image pre-training.
\newblock \emph{ICCV}, 2023.

\bibitem[Zhou et~al.(2022)Zhou, Lin, Zheng, Li, and Yang]{zhou2022not}
Zhou, J., Lin, Z., Zheng, Y., Li, J., and Yang, Z.
\newblock Not all tasks are born equal: Understanding zero-shot generalization.
\newblock In \emph{ICLR}, 2022.

\end{thebibliography}
\bibliographystyle{icml2025}
\newpage
\appendix
\onecolumn

\section{Theoretical properties of \texttt{Iso-C}}

In this Appendix, we discuss the theoretical properties of \texttt{Iso-C} by explicitly showing the connection between spectral skewness and the increased subspace dimensionality $k_{\text{M}}$ in the merged model achieved by \texttt{Iso-C}, which leads to a higher Subspace Alignment Ratio (SAR). Moreover, we explain why the increased SAR reduces inter-task interference. Finally, we highlight the limitations of $\texttt{Iso-C}$ that lead to the development of $\texttt{Iso-CTS}$. 
\subsection{Spectral skewness and the definition of $k_{\text{M}}$}
\label{apdx:skewness-kM-SAR}
In this Section, we show that the number of dominant components $k_{\text{M}}$ of the merged model $\Delta_{\text{M}}$ (see \cref{eq:rank}) is directly influenced by the skewness of its singular value spectrum. 
Using the singular value decomposition (SVD), let $\Delta_{\text{M}}=U\Sigma V^T$, where $\Sigma=\text{diag}(\sigma_1,\ldots,\sigma_r)$. By the definition of Frobenius norm:
$$\Vert \Delta_{\text{M}}\Vert_F^2=\sum_{i=1}^r\sigma_i^2,\quad\Vert\Delta_{\text{M}}-\Pi_{k,\text{M}}\Delta_{\text{M}}\Vert_F^2=\sum_{i=k+1}^r\sigma_i^2.$$  
Hence, the relative approximation error becomes:
$$\frac{\Vert\Delta_{\text{M}}-\Pi_{k,\text{M}}\Delta_{\text{M}}\Vert_F^2}{\Vert\Delta_{\text{M}}\Vert_F^2}=\frac{\sum_{i=k+1}^r\sigma_i^2}{\sum_{i=1}^r\sigma_i^2}.$$

Accordingly, $k_{\text{M}}$ can be defined in terms of singular values:

$$k_{\text{M}}=\text{min}\left\lbrace k:\frac{\sum_{i=k+1}^r\sigma_i^2}{\sum_{i=1}^r\sigma_i^2}\leq\epsilon^2\right\rbrace.$$
 
This formulation explicitly shows how the skewness of the spectrum $\{\sigma_i\}$ controls $k_\text{M}$. When $\Delta_{\text{M}}$ has a skewed spectrum (e.g. $\sigma_1^2 \gg \sum_{i=2}^r \sigma_i^2$), a small $k_{\text{M}}$ is sufficient to meet the error bound. This explains why Task Arithmetic $\Delta_{\mathrm{TA}}$ ($\beta=0$ in \Cref{fig:ta-iso-interpolation-spectrum}) -- which has a skewed spectrum -- yields a smaller $k_{\text{TA}}$ than \texttt{Iso-C}, whose flatter spectrum leads to a larger $k_{\texttt{Iso-C}}$. Therefore, expressing $k_{\text{M}}$ directly in terms of singular values highlights the link between the spectral skewness and subspace dimensionality.

\subsection{\texttt{Iso-C} increases Subspace Alignment Ratio (SAR)}
\label{apdx:iso-sar}
In this Section, we formally show how \texttt{Iso-C} increases Subspace Alignment Ratio (SAR) by expanding the effective subspace dimensionality of the merged model -- from $k_{\text{TA}}$ in Task Arithmetic to $k_{\texttt{Iso-C}}$ in \texttt{Iso-C}. 

The rank $k_{\text{M}}$ defines the effective rank of the subspace identified by the merged model and it is directly determined directly by its spectrum (as discussed \cref{apdx:skewness-kM-SAR}). Let $k_{\text{TA}}$ be the effective rank of $\Delta_{\text{TA}}$, and define
$$T=\lbrace u_1,..,u_{k_{\text{TA}}}\rbrace$$
as the orthonormal basis formed by those $k_{\text{TA}}$ singular vectors. Flattening the spectrum of $\Delta_{\mathrm{TA}}$ (\Cref{fig:ta-iso-interpolation-spectrum}), yields $\Delta_{\texttt{Iso-C}}$ with effective rank $k_{\texttt{Iso-C}}>k_{\mathrm{TA}}$ (as discussed in \cref{apdx:skewness-kM-SAR}). This flattening modifies only the singular values of TA, leaving the singular vectors unchanged. Therefore, the original subspace $T$ is contained within the larger subspace spanned by the top singular vectors of $\Delta_{\texttt{Iso-C}}$, defined as:
$$I=\lbrace u_1,..,u_{k_{\text{TA}}},..,u_{k_{\texttt{Iso-C}}}\rbrace.$$
Thus, by construction, we have $T\subset I$.

For simplicity, let $\Pi_T=\Pi_{k_{\text{TA}},\text{TA}}$ and $\Pi_I=\Pi_{k_{\texttt{Iso-C}},\text{\texttt{Iso-C}}}$ denote the projection operators onto the subspaces spanned by $T$ and $I$, respectively. Since $T\subset I$, for any matrix $\Delta_t$ it holds that:
\begin{equation}
\label{eq:disequality-sar}
\text{SAR}(\Delta_t,\Delta_{\mathrm{TA}}; k_{\text{TA}})=\frac{\Vert\Pi_T\Delta_t\Vert_F}{\Vert\Delta_t\Vert_F}\leq\frac{\Vert\Pi_I\Delta_t\Vert_F}{\Vert\Delta_t\Vert_F}=\text{SAR}(\Delta_t,\Delta_{\texttt{Iso-C}}; k_{\texttt{Iso-C}}),
\end{equation}

This inequality holds because by definition:
$$\frac{\Vert\Pi_T\Delta_t\Vert_F^2}{\Vert\Delta_t\Vert_F^2}=\frac{\sum_{i=1}^{k_{\mathrm{TA}}}\sum_j\langle u_i,\Delta_t^{(j)}\rangle^2}{\Vert\Delta_t\Vert_F^2}\leq\frac{\sum_{i=1}^{k_{\mathrm{TA}}}\sum_j\langle u_i, \Delta_t^{(j)}\rangle^2+\sum_{i=k_{\mathrm{TA}}+1}^{k_{\texttt{Iso-C}}}\sum_j\langle u_i,\Delta_t^{(j)}\rangle^2}{\Vert\Delta_t\Vert_F^2}=\frac{\Vert\Pi_I\Delta_t\Vert^2_F}{\Vert\Delta_t\Vert^2_F},$$
where $\Delta_t^{(j)}$ denotes the $j$-th column of $\Delta_t$. 

The equality in~\cref{eq:disequality-sar} holds only if the additional vectors added to the basis $T$ -- that is $\lbrace u_{k_{\mathrm{TA}}+1},\ldots,u_{k_{\texttt{Iso-C}}}\rbrace$ -- are orthogonal to each $\Delta^{(j)}_t$ or, equivalently, if they do not intersect the column space of $\Delta_t$ (i.e. its left singular vectors).

Hence, in general a lower $k_\mathrm{M}$ yields smaller or equal SAR than a larger $k_\mathrm{M}$. However, our empirical findings show that enriching the basis $T$ with singular vectors corresponding to smaller singular values in original task arithmetic spectrum (i.e. $\lbrace u_{k_{\mathrm{TA}}+1},\ldots,u_{k_{\texttt{Iso-C}}}\rbrace$) consistently increases the alignment ratio (\Cref{fig:ta-iso-interpolation-ar-vs-beta}), implying that these vectors are relevant for representing each task matrix $\Delta_t$ and not orthogonal to its left singular vectors. 

This analysis formally supports the claim that increasing the effective rank $k_\mathrm{M}$ of the merged matrix -- achieved by spectrum flattening in \texttt{Iso-C} -- leads to a higher Subspace Alignment Ratio.

\subsection{\texttt{Iso-C} mitigates inter-task interference}
\label{apdx:sar-interference}

\texttt{Iso-C} increases the Subspace Alignment Ratio (SAR), which quantifies how well the principal directions of a task matrix align with the principal directions of the merged model. In this Section, we demonstrate how a higher SAR contributes to mitigate inter-task interference by analyzing the relationship between subspace alignment and changes in internal activations following merging. Specifically, we define the interference as the degradation in a task’s internal representation due to merging —- that is, the deviation between the activations of the merged model and those of the corresponding single-task fine-tuned model. Intuitively, we can minimize the task interference by ensuring that the internal representations of task $j$ remain stable after merging.
 
Let $\theta_0$ be the pre-trained weights for a layer $l$. Define the task matrix $\Delta_j=\theta_j-\theta_0$ and the merged task matrix $\Delta_{\text{M}}$ for the layer $l$. Then, for an input $x_j^{(l)}$, we desire that the post-merging activation $h_j^{(l)}=(\theta_0+\alpha\Delta_{\text{M}})x_j^{(l)}$, with $\alpha$ chosen on a validation set, be close to the task-specific activation $\hat{h}_j^{(l)}=(\theta_0+\Delta_j)x_j^{(l)}$. Hence, we can quantify the interference as: 
\begin{equation}
\label{eq:interference}
   ||\hat{h}_j^{(l)}-h_j^{(l)}||=||(\Delta_j-\alpha\Delta_{\text{M}})x_j^{(l)}||\leq||\Delta_j-\alpha\Delta_{\text{M}}||\cdot||x_j^{(l)}||. 
\end{equation}

To show that the interference is lower when the Subspace Alignment Ratio (SAR) between $\Delta_j$ and $\Delta_{\text{M}}$ is higher, we decompose $\Delta_j$ into components aligned with and orthogonal to $\Delta_{\text{M}}$:
\begin{equation}
\label{eq:orthogonal-decomposition}
\Delta_j=\Delta_j^{||}+\Delta_j^{\perp} \quad \mbox{ where } \quad \Delta_j^{||}=\Pi_{k_\text{M},\text{M}}\Delta_j, \quad \Delta_j^{\perp}=(I-\Pi_{k_\text{M},\text{M}})\Delta_j,
\end{equation}
and $\Pi_{k_{\text{M}}, \text{M}}$ is the projection matrix onto the subspace spanned by the top $k_{\text{M}}$ left-singular vectors of $\Delta_{\text{M}}$ (see Equation~\eqref{eq:rank} for the definition of $k_{\text{M}}$). The Subspace Alignment Ratio is then:
\begin{equation}
\label{eq:rewrite-sar}
      \text{SAR} (\Delta_{j}, \Delta_{\text{M}}; k_{\text{M}}) = \frac{||\Pi_{k_{\text{M}}, \text{M}}\Delta_{j}||_{F}}{||\Delta_{j} ||_{F}} = \frac{||\Delta_j^{||}||_F}{||\Delta_j^{||}+\Delta_j^{\perp}||_F}.
\end{equation}
Similarly, decomposing $\Delta_\text{M}$ into $\Delta_{\text{M}}^{||}$ and $\Delta_{\text{M}}^{\perp}$ and substituting \cref{eq:orthogonal-decomposition} in \cref{eq:interference}, the interference becomes:  
\begin{equation}
\label{eq:rewrite-interference}
||\Delta_j-\alpha\Delta_{\text{M}}||=||\Delta_j^{||}-\alpha\Delta^{||}_{\text{M}}+\Delta_j^{\perp}-\alpha\Delta^{\perp}_{\text{M}}||\approx||\Delta_j^{||}-\alpha\Delta^{||}_{\text{M}}+\Delta_j^{\perp}||,
\end{equation}
since $k_{\text{M}}$ minimizes the approximation error of $\Delta_{\text{M}}$, leading to $||\Delta^{\perp}_{\text{M}}||\approx0$.

If the SAR defined in \cref{eq:rewrite-sar} is close to 1, then $||\Delta_j^{\perp}||$ is small, so the interference in \cref{eq:rewrite-interference} mainly depends on $||\Delta_j^{||}-\alpha\Delta^{||}_M||$. Conversely, if SAR is near zero, the large orthogonal component $\Delta_j^{\perp}$ increases the overall interference, regardless of the choice of $\alpha$. Even with an optimal $\alpha$ chosen via validation, interference cannot be reduced below the norm of the orthogonal component.

\texttt{Iso-C} increases the SAR of $\Delta_t$ with the merged model — bringing it close to 1, as shown in the paper — by flattening the singular values. Thus, the optimal $\alpha$ can adjust the merged model such that interference is minimized. In contrast, Task Arithmetic (TA), with SAR varying across tasks, exhibits interference that cannot be reduced below the norm of the orthogonal component. We experimentally evaluate that the interference is lower for \texttt{Iso-C} than TA in \Cref{apdx:exp-interference}.

\subsection{Limitations of \texttt{Iso-C} that motivate \texttt{Iso-CTS}}
\label{apdx:iso-c-limitations}

This Section details the limitations of \texttt{Iso-C} that motivate the development of \texttt{Iso-CTS}. Specifically, \texttt{Iso-C} relies on the singular vectors obtained through Task Arithmetic to perform model merging. As a result, it tends to underrepresent tasks whose dominant directions have lower intensity compared to the majority, particularly when those directions are orthogonal to the shared (common) directions. This limitation becomes increasingly pronounced as the number and diversity of tasks increase (see \Cref{sec:isocts}).

To make this limitation explicit, we formalize the computation -- via SVD -- of the first left singular vector in Task Arithmetic, used by \texttt{Iso-C}, as the variance maximization problem:
$$u_1=\arg\max_{||u|\vert=1}||\Delta_{\text{TA}}^{\top}u||^2=u^{\top}\left(\sum_{t=1}^T\Delta_t\Delta_t^{\top}\right)u+u^{\top}(\sum_{{t,s=1,t\neq s}}^T\Delta_t\Delta_s^{\top})u$$
If a particular task $\Delta_j$ has dominant directions with significantly lower intensity compared to the other tasks (i.e. lower Frobenius Norm), then its individual contributions $\Delta_j \Delta_j^{\top}$ to the total variance becomes smaller. Similarly, cross terms involving $\Delta_j$ will also be comparatively small.  Therefore, task $j$ explicitly contributes less to the maximized variance captured by the first principal singular direction.

Moreover, if the directions of $\Delta_j$ are orthogonal or nearly orthogonal to $u_1$, (i.e. $u_1^{\top}\Delta_j=0$), task $j$ contributes minimally or not at all along this principal direction. Similar considerations apply to subsequent singular vectors $u_2, \ldots u_k$, defining the common subspace. Finally, as the number of tasks $T$ increases and tasks become more diverse, it becomes increasingly likely that tasks with distinct but smaller-magnitude directions will be underrepresented or absent in the dominant singular directions identified by the task arithmetic decomposition. 

The goal of \texttt{Iso-CTS} is to address this limitation by incorporating orthogonal directions that are overlooked by the Task Arithmetic spectrum. This strategy yields the greatest improvements in settings with a large number of diverse tasks, as shown in our experimental results.

\section{Computational complexity analysis} 
\label{apdx:computational-complexity}
In this Section, we analyze the computational complexity of \texttt{Iso-C} and \texttt{Iso-CTS} and compare it with that of our main competitor, TSV-M~\cite{tsv}.

Let $\Delta_t \in \mathbb{R}^{n\times n}$, and let $T$ and $L$ be the number of tasks and network layers, respectively. For simplicity, assume that each layer consists of a single squared $n \times n$ matrix.

In our analysis, we focus on the number of SVD performed by each algorithm, as this is by far the most costly component of each algorithm. The complexity of a single SVD on $\Delta_t \in \mathbb{R}^{n\times n}$ is equal to $\mathcal{O}(n^3)$~\cite{vasudevan2017hierarchical}. Below, we detail the total computational complexity for each merging method:
\begin{itemize}
    \item \texttt{Iso-C} performs a single SVD on $\Delta_{\text{TA}}$ per layer, with total complexity:  
    \begin{equation*}
      \mathcal{O}(\texttt{Iso-C})=\mathcal{O}(Ln^3)  
    \end{equation*}
    \item \texttt{Iso-CTS} performs: 
    \begin{itemize}
        \item One SVD on $\Delta_{\text{TA}}$ per layer (lines 2-3, \Cref{alg:method_common_and_task_specific}) with complexity $\mathcal{O}(Ln^3)$
        \item One SVD on each $\Delta_{t}$, for all $T$ tasks and each of the $L$ layers (line 5, \Cref{alg:method_common_and_task_specific}), with  complexity $\mathcal{O}(TLn^3)$
        \item Two SVDs on two matrices $U_{*},V_{*} \in \mathbb{R}^{n \times n}$ per layer (line 11, \Cref{alg:method_common_and_task_specific}), with complexity of $\mathcal{O}(2Ln^3)$.
    \end{itemize}
    
    Therefore, the total complexity equals:
    \begin{equation*}
           \mathcal{O}(\texttt{Iso-CTS})=\mathcal{O}(Ln^3 + TLn^3 + 2Ln^3)=\mathcal{O}((T+3)Ln^3)=\mathcal{O}(TLn^3) 
    \end{equation*}

    \item TSV-M~\cite{tsv} performs: 
    \begin{itemize}
        \item $T$ SVDs per layer on each task matrix (line 1, Alg. 1 from~\citet{tsv}): $\mathcal{O}(T L n^3)$ \
        \item  Two additional SVDs per layer (lines 10-11, Alg.1 from~\citet{tsv}): $\mathcal{O}(2 L n^3)$
    \end{itemize}
     Yielding the total complexity: 
    $$\mathcal{O}(\mathrm{TSV}) = \mathcal{O}(TLn^3 + 2Ln^3) = \mathcal{O}((T+2)Ln^3) = \mathcal{O}(TLn^3)$$
\end{itemize}

While \texttt{Iso-CTS} and TSV-M share the same asymptotic complexity, \texttt{Iso-CTS} incurs slightly more overhead due to the SVD on $\Delta_{\text{TA}}$ (lines 2-3, \Cref{alg:method_common_and_task_specific}). Both methods can be further optimized by computing Truncated SVDs for \texttt{Iso-CTS} and TSV-M, since only a few components are retained. This reduces the complexity for both approaches. \texttt{Iso-C} is the most computationally efficient algorithm -- its complexity is constant with respect to number of task $T$.

\section{Experimental details}
In this Appendix, we provide the dataset and implementation details used to carry out the experiments presented in the paper.

\subsection{Datasets}
\label{apdx:datasets}

The 8-dataset benchmark consists of: Cars~\cite{cars}, DTD~\cite{dtd}, EuroSAT~\cite{eurosat}, GTSRB~\cite{gtsrb}, MNIST~\cite{MNIST}, RESISC45~\cite{cheng2017remote}, SUN397~\cite{sun397}, and SVHN~\cite{svhn}.

The 14-dataset benchmark builds on the preceding one, incorporating six additional datasets: CIFAR100~\cite{krizhevsky_learning_nodate}, STL10~\cite{coates_analysis_2011}, Flowers102~\cite{nilsback_automated_2008}, OxfordIIITPet~\cite{parkhi_cats_2012}, PCAM~\cite{veeling_rotation_2018}, and FER2013~\cite{goodfellow_challenges_2013}.

Finally, the 20-dataset benchmark includes the preceding 14 plus the following six: EMNIST~\cite{cohen_emnist_2017}, CIFAR10~\cite{krizhevsky_learning_nodate}, Food101~\cite{bossard_food-101_2014}, FashionMNIST~\cite{xiao_fashion-mnist_2017}, RenderedSST2~\cite{socher_recursive_nodate}, and KMNIST~\cite{clanuwat_deep_2018}. 

\subsection{Implementation details}

Our method relies on SVD, which is defined for two-dimensional matrices $\Delta \in \mathbb{R}^{m\times n}$. However, some weights of the neural networks are represented by vectors $\delta \in \mathbb{R}^{n}$, e.g. bias vectors and parameters of layer normalization~\cite{ba2016layer}. Therefore, following~\citet{tsv}, we apply simple averaging to combine these parameters.

\section{Additional experiments}
In this Appendix, we present additional experiments that complement the main paper, including comparisons with new vision baselines using Task Arithmetic model checkpoints~\cite{ilharco2023task} for evaluation. We empirically assess the reduced interference of \texttt{Iso-C} compared to Task Arithmetic and analyze the impact of the scaling factor $\alpha$ on our approaches. Finally, we present an ablation study showing what happens when spectrum flattening is applied to each task model individually.

\subsection{Additional vision baselines}
\label{apdx:additional-baselines}

In this Section we provide results with additional methods: Fisher Merging~\cite{matena2021merging}, RegMean~\cite{jin2023dataless}, PCB~\cite{guodong24neurips}, MaTS~\cite{tam2023merging} and CART~\cite{lee2025adarankadaptiverankpruning}. These methods were originally evaluated on checkpoints from Task Arithmetic~\cite{ilharco2023task} provided for 8 tasks on ViT-B/32 and ViT-L/14. We follow this experimental protocol with \texttt{Iso-C} and \texttt{Iso-CTS} and present the results in \Cref{tab:multitask_acc_TA_ckpts}. \texttt{Iso-CTS} sill achieves state-of-the-art performance followed by \texttt{Iso-C}. Note that the results differ from \Cref{tab:multitask_acc} where we used checkpoints from Consensus Merging~\cite{wang2024localizing}.

\begin{table}[t!]
  \centering
  \caption{Additional baselines for merging ViT-B/32 and ViT-L/14 on 8 tasks. We report absolute accuracy.}
  \label{tab:multitask_acc_TA_ckpts}
  \vspace{0.03in}
  \resizebox{0.35\textwidth}{!}{
    \begin{tabular}{l|cc}
      \toprule
      \textbf{Method} & \textbf{ViT-B/32} & \textbf{ViT-L/14} \\
      \midrule
      Zero-shot & 48.3 & 64.7 \\
      Fine-tuned & 90.5 & 94.2 \\
      \midrule
      Task Arithmetic & 70.5 & 84.6 \\
      Fisher Merging & 68.3 & 83.7 \\
      RegMean & 71.8 & 82.2 \\
      PCB & 76.3 & 87.5 \\
      MaTS & 82.6 & 90.2 \\
      CART & 83.0 & 90.8 \\
      \midrule
      \rowcolor{isoc!20}
      \textbf{\texttt{Iso-C} (Ours)} & \underline{84.1} & \underline{92.5} \\
      \rowcolor{isocts!25}
      \textbf{\texttt{Iso-CTS} (Ours)} & \textbf{84.3} & \textbf{93.0} \\
      \bottomrule
    \end{tabular}
  }
\end{table}

\newpage 
\subsection{Interference quantification}
\label{apdx:exp-interference}

\begin{figure}[t]
    \centering
    \includegraphics[width=0.65\linewidth]{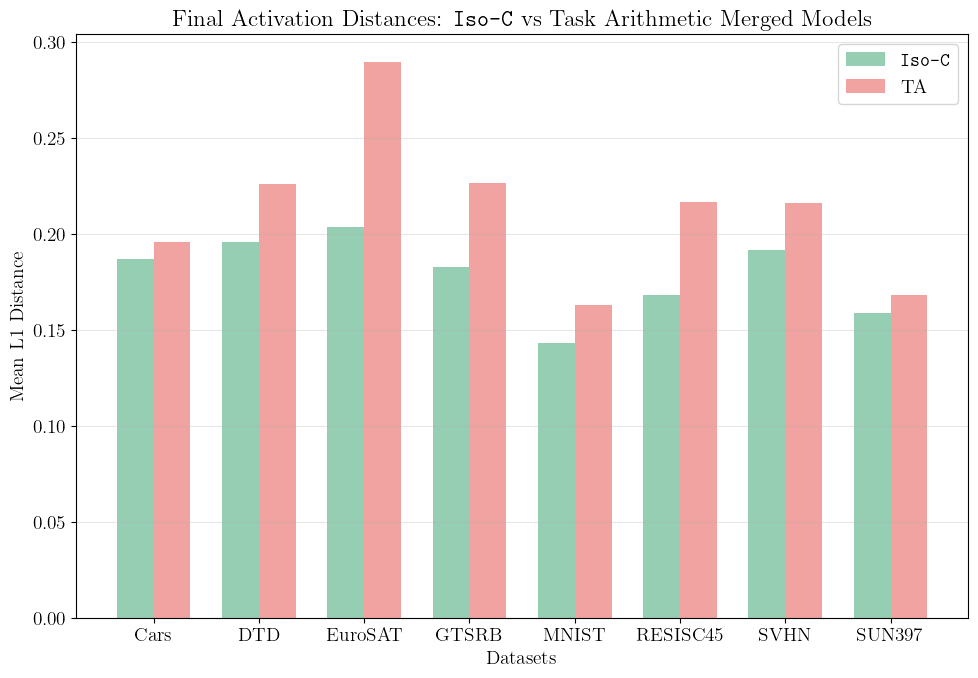}
    \vspace{-0.1in}
    \caption{Mean L1 distance between the final embeddings of task-specific models and the merged one for \texttt{Iso-C} and TA. We used ViT-B/16 model.}
    \label{fig:l1-distances}
\end{figure}

In this Section, we experimentally show that merging interference (defined in \Cref{apdx:sar-interference}) is lower when merging is performed with \texttt{Iso-C} than with TA. Following~\citet{yang2024representation}, we measure the interference as L1 distance between the final embeddings of task-specific models and merged one. 
In \Cref{fig:l1-distances} we present the results for merging 8 tasks on  ViT-B/16. We observe that the interference is lower for \texttt{Iso-C} than for TA highlighting the effectiveness of \texttt{Iso-C} in reducing interference when merging models.

\subsection{Selection of scaling coefficient $\alpha$}

\begin{table}[t]
    \begin{center}
    \caption{Optimal $\alpha$ value chosen on a held-out validation set for different model types and numbers of tasks for \texttt{Iso-C} and \texttt{Iso-CTS}.}
    \vspace{0.05in}
    \scalebox{0.85}{
    \begin{tabular}{c|c|c|c|c}
        \toprule
        \textbf{Method} & \textbf{Model} & \textbf{8 tasks} & \textbf{14 tasks} & \textbf{20 tasks} \\
        \midrule
        \multirow{3}{*}{\texttt{Iso-C}} & ViT/32-B & $1.30$ & $1.00$ & $0.90$ \\
                                  & ViT/16-B & $1.40$ & $1.00$ & $0.80$ \\
                                  & ViT/14-L & $1.50$ & $1.30$ & $1.00$ \\
        \midrule
        \multirow{3}{*}{\texttt{Iso-CTS}} & ViT/32-B & $1.50$ & $1.20$ & $1.10$ \\
                                  & ViT/16-B & $1.60$ & $1.20$ & $1.10$ \\
                                  & ViT/14-L & $1.90$ & $1.50$ & $1.20$ \\
        \bottomrule
    \end{tabular}
    }
    \label{tab:alpha_table}
    \end{center}
\end{table}

\begin{figure}[t]
    \includegraphics[width=\linewidth]{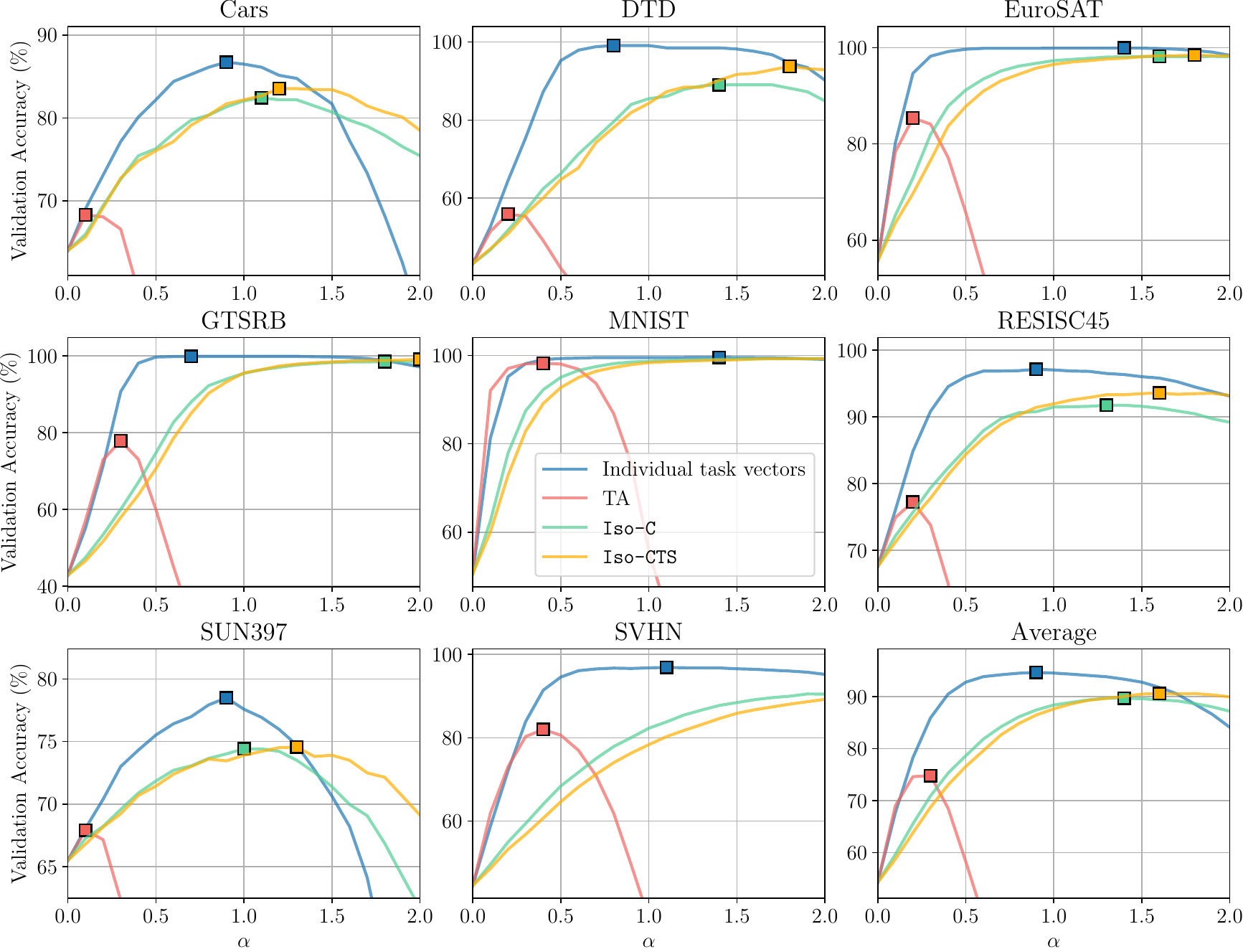}
    \caption{TA is sensitive to the selection of $\alpha$, while both \texttt{Iso-C} and \texttt{Iso-CTS} are more robust to $\alpha$ selection, resembling the task-specific models. The $\alpha$ is chosen based on the best average performance on the validation set across tasks. The bottom right subplot denotes the optimal $\alpha$ for each method (Eq. \eqref{eq:ta}, Eq.  \eqref{eq:isoc-alpha} and Eq. \eqref{eq:isocts-alpha}). The model is ViT-B/16.}
    \label{fig:acc-vs-alpha}
\end{figure}

In \cref{fig:acc-vs-alpha}, we present the relationship between the validation accuracy and scaling factor $\alpha$. We observe that TA is very sensitive to the selection of $\alpha$, which potentially may require a more fine-grained search. On the other hand, both \texttt{Iso-C} and \texttt{Iso-CTS} are more robust to $\alpha$ selection, resembling the task-specific models. For reproducibility, In Table \ref{tab:alpha_table}, we provide the optimal $\alpha$ value chosen on the held-out validation set for each model and number of tasks.

\subsection{Importance of isotropic scaling in \texttt{Iso-CTS}}

\begin{figure}[t]
    \centering
    \includegraphics[width=0.7\linewidth]{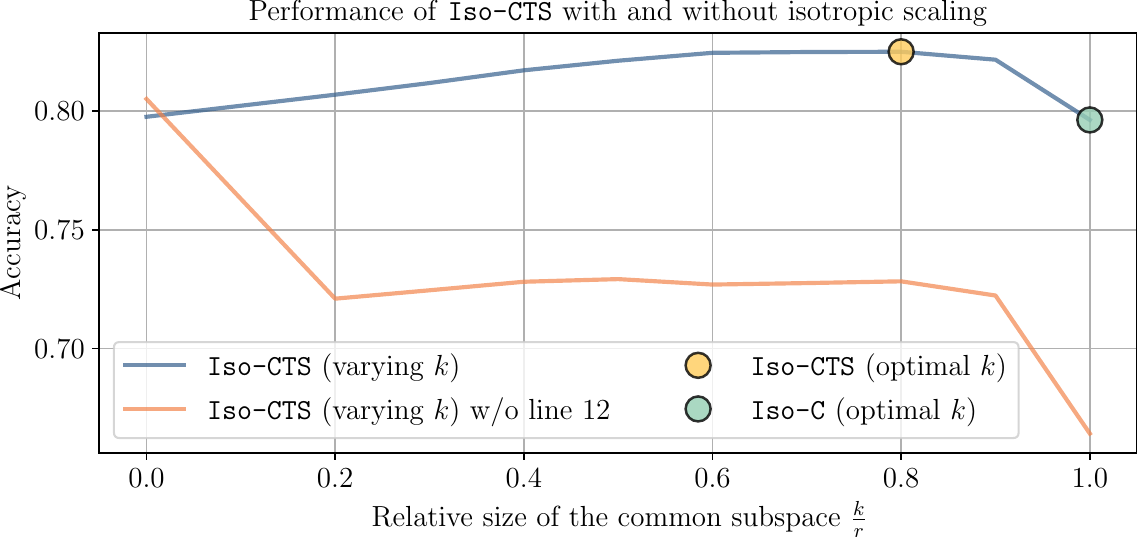}
    \caption{Performance of \texttt{Iso-CTS} with and without isotropic scaling (Eq.\ref{eq:iso-cts}). Isotropic scaling is a crucial component of \texttt{Iso-CTS}. Results for merging 20 tasks with ViT-B/16.}
    \label{fig:frac-common-space-w-wo-iso}
\end{figure}

In this Section we ablate the need for isotropic scaling in \texttt{Iso-CTS}.
We present the comparison of the performance of \texttt{Iso-CTS} with and without isotropic scaling (Equation \ref{eq:iso-cts}) in \Cref{fig:frac-common-space-w-wo-iso}. We observe that isotropic scaling is indeed a crucial component of \texttt{Iso-CTS} as long as common subspace exists. When only task-specific subspaces are in use ($\frac{k}{r}=0$), isotropic scaling does not make a significant difference.
However, the design in \Cref{alg:method_common_and_task_specific} also plays an important role, especially when the number of merged models increases, leading to up to 2.8\% improvement over \texttt{Iso-C} on 20 tasks (see \Cref{tab:multitask_acc}).

\begin{figure}[t]
    \centering
    \includegraphics[width=0.5\linewidth]{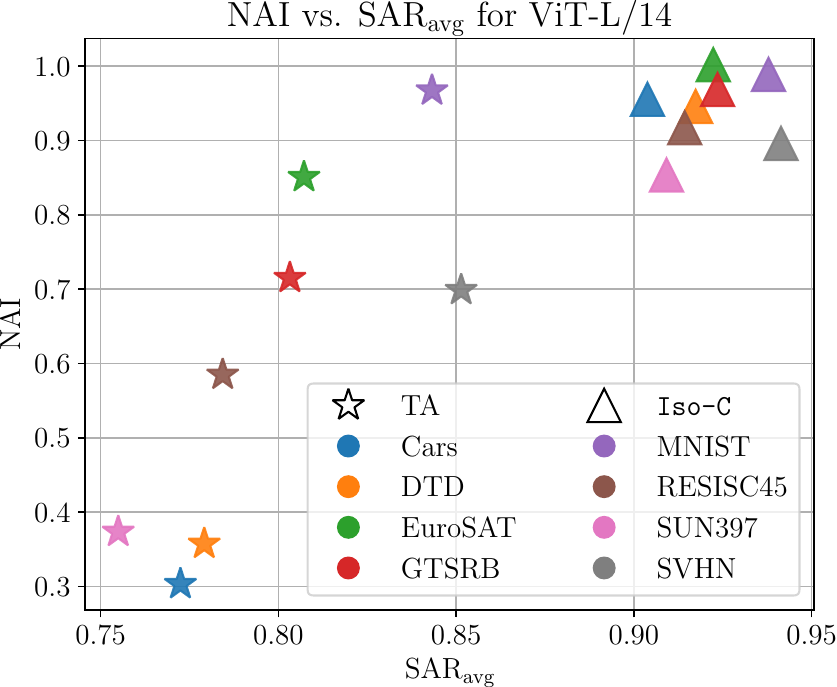}
    \caption{Normalized Accuracy Improvement (NAI) vs. Average Subspace Alignment Ratio ($\text{SAR}_\text{avg}$) for ViT-L/14.}
    \label{fig:nai-vs-ar-ViT-L-14}
\end{figure}

\subsection{Applying \texttt{Iso} to individual task matrices}

\begin{figure}[t]
    \centering
    \includegraphics[width=0.8\linewidth]{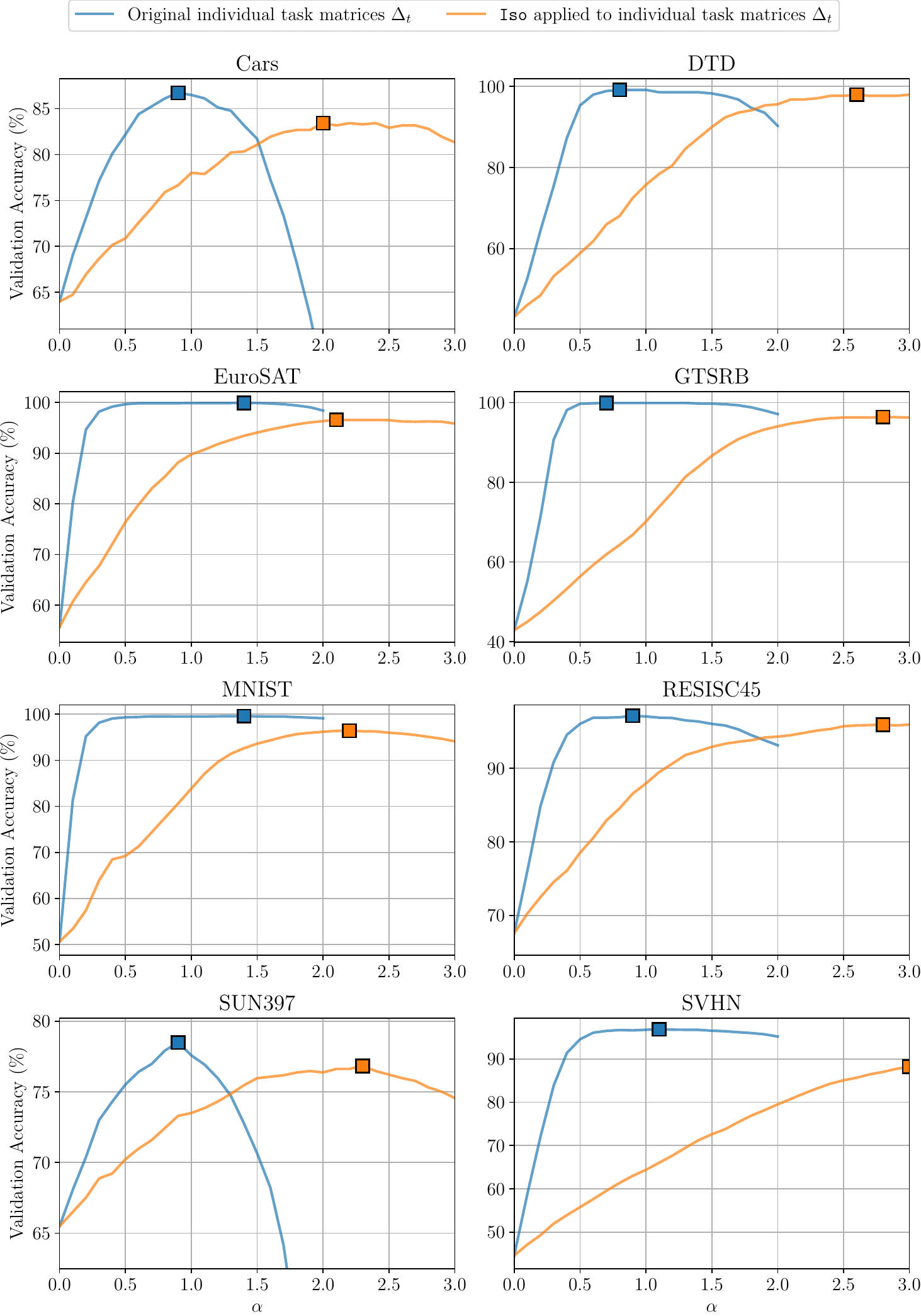}
    \caption{Validation Accuracy while scaling task matrices with $\alpha$ coefficient (Eq.~\eqref{eq:ta} applied for a single task). We observe a performance gap between the accuracy of original and modified models for the optimal values of $\alpha$ (denoted by square).}
    \label{fig:ind-orig-vs-ind-iso}
\end{figure}

Flattening the skewed spectrum of singular values significantly improves the performance of the merged model, as demonstrated in \cref{sec:analysis}. One may wonder if this operation might also be an effective strategy for improving single-task models. \cref{fig:ind-orig-vs-ind-iso} presents the performance of task-specific models in their original form along with their modified versions with singular value spectra of their task matrices flattened (which is equivalent to performing \texttt{Iso-C} for a single model). We observe a 3.3\% drop in average performance across tasks. Therefore, the reason for the success of \texttt{Iso-C} lies in its ability to mitigate the negative effects of summing task matrices, not in inadvertently improving the original individual task matrices.

\section{Additional visualizations}
In this Appendix, we provide additional visualizations that could not be included in the main paper due to space constraints. These include the spectra of task matrices, the Subspace Alignment Ratio per layer, and the correlation between Normalized Accuracy Improvement and Subspace Alignment Ratio when using the larger ViT-L/14 model.

\subsection{Visualization of task matrix spectra}

\begin{figure}[t]
    \centering
    \includegraphics[width=\linewidth]{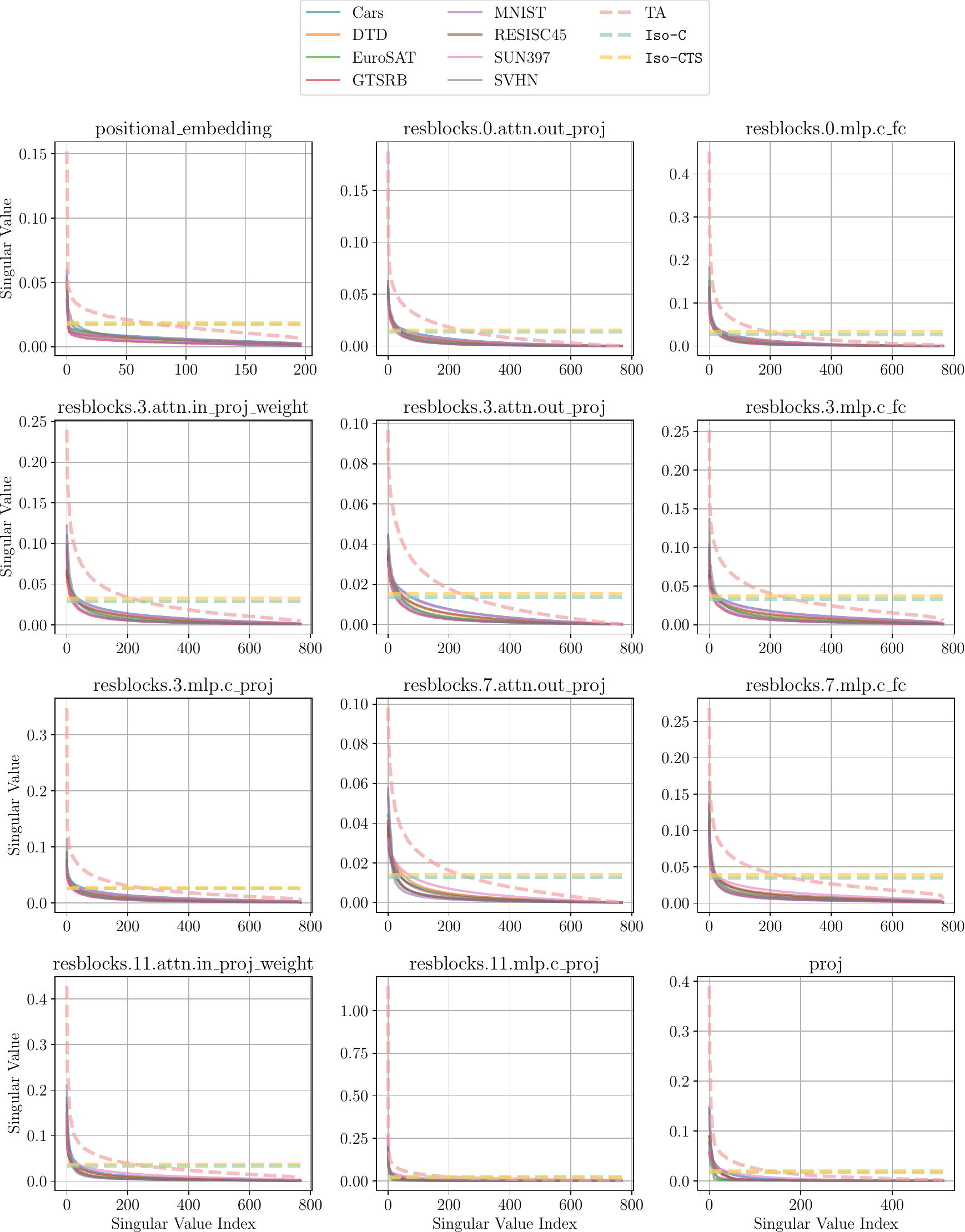}
    \caption{Visualization of singular value spectra of different task matrices for different types of layers in ViT/B-16.}
    \label{fig:combined-spectra}
\end{figure}

When visualizing spectra of singular values of task matrices (\cref{fig:teaser} and \cref{fig:ta-iso-interpolation-spectrum}), we selected an output projection matrix $W^O$ from layer $\ell=4$ of ViT/B-16 as an illustrative example. In~\cref{fig:combined-spectra}, we present spectra across a variety of layers of ViT/B-16 for the task matrices of task-specific models, TA, \texttt{Iso-C} and \texttt{Iso-CTS}.

\subsection{Visualization of per layer Subspace Alignment Ratio}

\begin{figure}[t]
    \centering
    \includegraphics[width=0.9\linewidth]{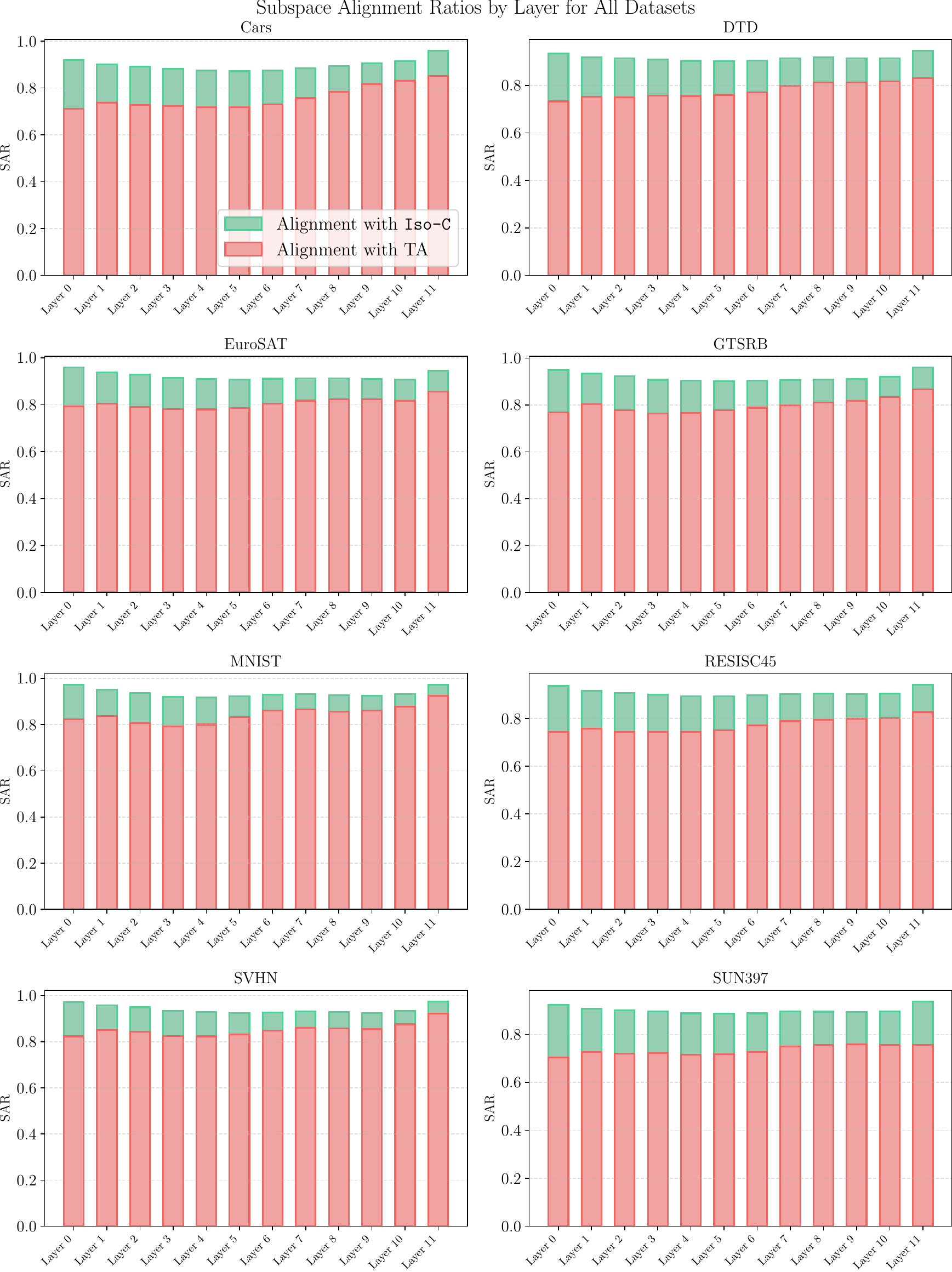}
    \caption{Per layer Subspace Alignment Ratio between fine-tuned and merged weight matrices for ViT-B/16.}
    \label{fig:alignment-by-layer}
\end{figure}

In \Cref{fig:motivation_nai_vs_ar} and \Cref{fig:ta-iso-interpolation-ar-vs-beta} in the main paper, we presented $\text{SAR}_\text{avg}$ -- Subspace Alignment Ratio averaged across all the task matrices.  
In this Section, we present SAR at different depths of the model. Specifically,  we calculate SAR between fine-tuned and merged weight matrices and an average of all the matrices for a given layer of the ViT-B/16 model. We present the results in \Cref{fig:alignment-by-layer}. We observe that the alignment is higher for \texttt{Iso-C} across all layers of the vision transformer. One may expect early layers to be more aligned but we find that for both approaches the alignment is similar across the layers.

\subsection{Visualization of Normalized Accuracy Improvement versus Subspace Alignment Ratio for ViT-L/14}

In \Cref{fig:nai-vs-ar-ViT-L-14} we replicate the experiment from \Cref{fig:motivation_nai_vs_ar} (conducted on ViT-B/16) on ViT-L/14. The observations from the main paper hold -- Normalized Accuracy Improvement strongly correlates with average Subspace Alignment Ratio, and increasing $\mathrm{SAR}_{\mathrm{avg}}$ via merging with \texttt{Iso-C} leads to better performance.

\end{document}